%% file: main.tex
\documentclass[journal]{IEEEtran}
\usepackage{amsmath,amsfonts}
\usepackage{algorithmic}
\usepackage{algorithm}
\usepackage{array}
\usepackage{textcomp}
\usepackage{stfloats}
\usepackage{url}
\usepackage{verbatim}
\usepackage{graphicx}
\usepackage{cite}

\usepackage[dvipsnames]{xcolor}
\input{preamble}

\newcommand\modified[0]{}
\usepackage[normalem]{ulem}
\usepackage{threeparttable}
\usepackage{adjustbox}
\usepackage{wrapfig}

\usepackage{tabularx,booktabs}
\newcolumntype{Y}{>{\centering\arraybackslash}X}

\definecolor{cvprblue}{rgb}{0.21,0.49,0.74}
\usepackage[pagebackref,breaklinks,colorlinks,citecolor=cvprblue]{hyperref}
\usepackage{xcolor} 
\usepackage{maths}

\usepackage{array} 
\usepackage{subcaption}
\usepackage{multirow}
\usepackage{soul}       
\usepackage{amssymb}

\newcommand*{\affaddr}[1]{#1} 
\newcommand*{\affmark}[1][*]{\textsuperscript{#1}}

\def\onedot{.}

\def\etal{\emph{et al}\onedot}

\def\maketitlesupplementary
   {
   \newpage
       \twocolumn[
        \centering
        \Large
        \textbf{\thetitle}\\
        \vspace{0.5em}Supplementary Material \\
        \vspace{1.0em}
       ] 
   }

\let\titleold\title
\renewcommand{\title}[1]{\titleold{#1}\newcommand{\thetitle}{#1}}
\def\maketitlesupplementary
   {
   \newpage
       \twocolumn[
        \centering
        \Large
        \textbf{\thetitle}\\
        \vspace{0.5em}Supplementary Material \\
        \vspace{1.0em}
       ] 
   }

\author{
Guillermo Gomez-Trenado\affmark[1]\affmark[2], Pablo Mesejo\affmark[1], Oscar Cordón\affmark[1], and Stéphane Lathuilière\affmark[3]\\
\small{\affaddr{\affmark[1]University of Granada and DaSCI Research Institute, Granada 18014, Spain}}\\
\small{\affaddr{\affmark[2]Panacea Cooperative Research, Ponferrada 24402, Spain}}\\
\small{\affaddr{\affmark[3]Inria at University Grenoble Alpes, Montbonnot-Saint-Martin, 38330, France}}\\

}

\begin{document}

\title{Don't Forget your Inverse DDIM for Image Editing}

\maketitle

\IEEEpubid{%
  \makebox[\columnwidth]{\hfill Corresponding author: Guillermo Gomez‑Trenado (guillermogomez@ugr.es) \hfill}%
  \hspace{\columnsep}%
  \makebox[\columnwidth]{}%
}
\IEEEpubidadjcol

\input{sec/0_abstract}    

\begin{IEEEkeywords}
Diffusion Model, Image Editing, Image Generation, Image Synthesis, Prompt-Based Editing, Generative Models
\end{IEEEkeywords}

\input{sec/1_intro}

\input{sec/2_related}

\input{sec/3_method}
\input{sec/4_experiments} 
\input{sec/5_limitations}

\input{sec/6_conclusions}
\input{sec/7_acknowledgements}

\bibliographystyle{IEEEtran}
\bibliography{main}


\input{sec/X_suppl}

\end{document}

%% file: preamble.tex
\usepackage[dvipsnames]{xcolor}


%% file: sec/0_abstract.tex
\begin{abstract}

    The field of text-to-image generation has undergone significant advancements with the introduction of diffusion models. {\modified{}Nevertheless}, the challenge of editing real images persists, as most methods are {\modified{}either computationally intensive or produce poor reconstructions}. {\modified{}This paper introduces SAGE} (Self-Attention Guidance for image Editing) —a novel technique leveraging pre-trained diffusion models for image editing. SAGE builds upon the DDIM algorithm and incorporates a novel guidance mechanism utilizing the self-attention layers of the diffusion U-Net. This mechanism computes a reconstruction objective based on attention maps generated during the inverse DDIM process, enabling efficient reconstruction of unedited regions without the need to precisely reconstruct the entire input image. Thus, SAGE directly addresses the key challenges in image editing. The superiority of SAGE over other methods is demonstrated through quantitative and qualitative evaluations and confirmed by a statistically validated comprehensive user study, in which all 47 surveyed users preferred SAGE over competing methods. Additionally, SAGE ranks as the top-performing method in seven out of 10 quantitative analyses and secures second and third places in the remaining three.
    
\end{abstract}

%% file: sec/1_intro.tex
\section{Introduction}
\label{sec:intro}

The advancements in text-guided image synthesis through diffusion models have garnered considerable attention due to their ability to achieve remarkable realism and diversity \cite{saharia2022imagen,rombach2022latent}. These large-scale models enable image generation from text prompts and have unlocked a new level of creativity. 
As a result, research is intensifying around the applications of these models to manipulate the underlying distributions of images for editing purposes. One striking innovation is the possibility of editing images through intuitive text prompts, giving users the  power to modify images without professional editing skills. {\modified{}This paper focuses} on the prompt-based image editing task as formulated in \cite{mokady2023null}: a user provides an image alongside its textual description. Then, by changing the meaning of the sentence, the user can instruct the model to add, omit, change, or enhance image elements (See Fig.~\ref{fig:teaser} for example). The models implicitly determine which areas of the input image are irrelevant to the target task and should be reconstructed, and which areas require effective modification while preserving the relevant identity and geometry.

\begin{figure*}[ht]
    \centering
    \includegraphics[width=0.9\linewidth]{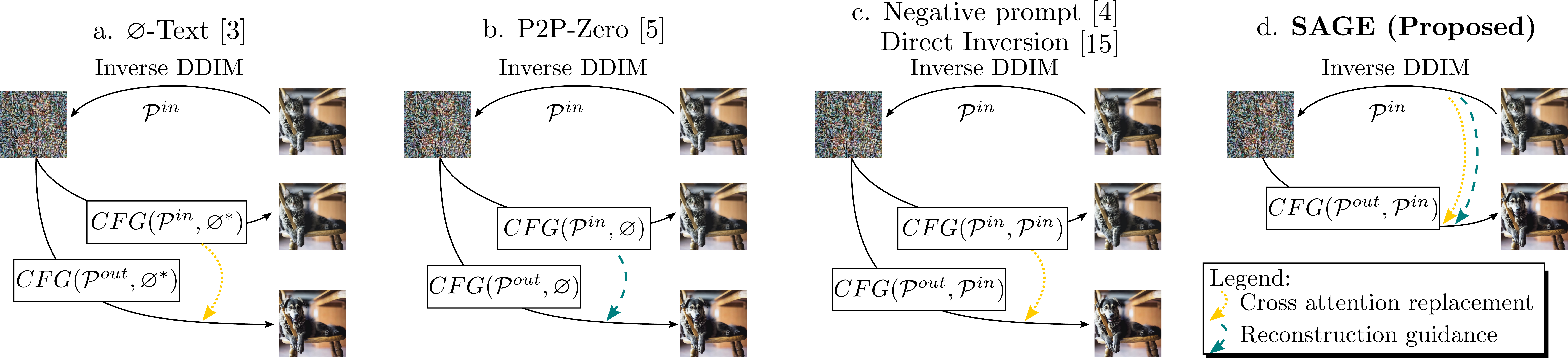}
    \caption{Comparative Analysis of Diffusion-Based Image Editing Techniques. {\modified{}This review contrasts} existing methodologies, which utilize Classifier-Free Guidance (CFG) \cite{ho2021classifier} with various combinations, including the pretrained null-prompt $\varnothing$, an optimized latent representation $\varnothing^*$, the descriptive prompt of the input image $\pin$, and the target editing prompt $\pout$. 
    }
      \label{fig:related}
    \end{figure*}

The state-of-the-art methods for the prompt-guided editing task require inversion of the target image (see Fig.~\ref{fig:related}). Although inversion processes have greatly improved within Generative Adversarial Networks, they remain a significant hurdle in diffusion models due to their iterative sampling process. Current techniques~\cite{mokady2023null} require repetitive optimization steps, resulting in high computational demands with even moderately-sized images (\(512 \times 512\)), taking upwards of a minute to process per image. Alternatives that reduce computational workload~\cite{mokady2023null,miyake2023negative,parmar2023zero} often compromise on reconstruction quality, resulting in undesired alterations of the input image.

\input{fig/teaser_figure.tex}

To address these challenges, {\modified{}this paper introduces} \method{}  (Self-Attention Guidance for Editing), a novel method that reconciles the requirements for computational efficiency and high-fidelity reconstruction, while affording flexible image editing capabilities. Our approach, akin to existing methodologies~\cite{mokady2023null,miyake2023negative,parmar2023zero}, leverages Denoising Diffusion Implicit Model (DDIM) \cite{song2021ddim} inversion. However, it uniquely exploits the intermediate self-attention and cross-attention maps internally computed by the diffusion model during the inverse DDIM process, enabling faithful reconstruction with minimal computational overhead. During the sampling process from random noise to the output image, {\modified{}our method benefits} from a synergistic application of Classifier-Free Guidance (CFG, see Sec~\ref{sec:self_attention_guidance}) alongside a novel reconstruction guidance mechanism based on self-attention reconstruction. This mechanism operates on the self-attention maps within the diffusion U-Net~\cite{ronneberger2015unet}, facilitating a fine trade-off between editing and preservation of the original image details.

In summary, our contribution is threefold:

\begin{itemize}
    \item {\modified{}Introduction of} a novel editing framework utilizing a pre-trained diffusion model that leverages intermediate noise vectors from the inverse DDIM process. This enables the reconstruction to be steered toward the input image, while allowing modifications aligned with textual prompts.
    \item {\modified{}Proposal of} a reconstruction guidance loss term that operates in the self-attention layers of the diffusion network. This term ensures high-fidelity reconstruction in regions unaffected by the editing process without adding major computational overhead.
    \item {\modified{}Experimental validation benchmarks our approach} against recent methods in the field. \method{} outperforms other methods not only in quantitative and qualitative evaluations but also in a comprehensive user study, where it was preferred in over 60\% of cases. Additionally, it ranked first in seven out of 10 quantitative assessments, securing second and third places in the remaining ones.
\end{itemize}

%% file: fig/teaser_figure.tex
  \begin{figure}[t]
    \footnotesize
    \small
    \centering
    \begin{tabular}{m{2.8cm}m{0.5cm}m{2.8cm}}
      \includegraphics[width=2.8cm, height=2.8cm]{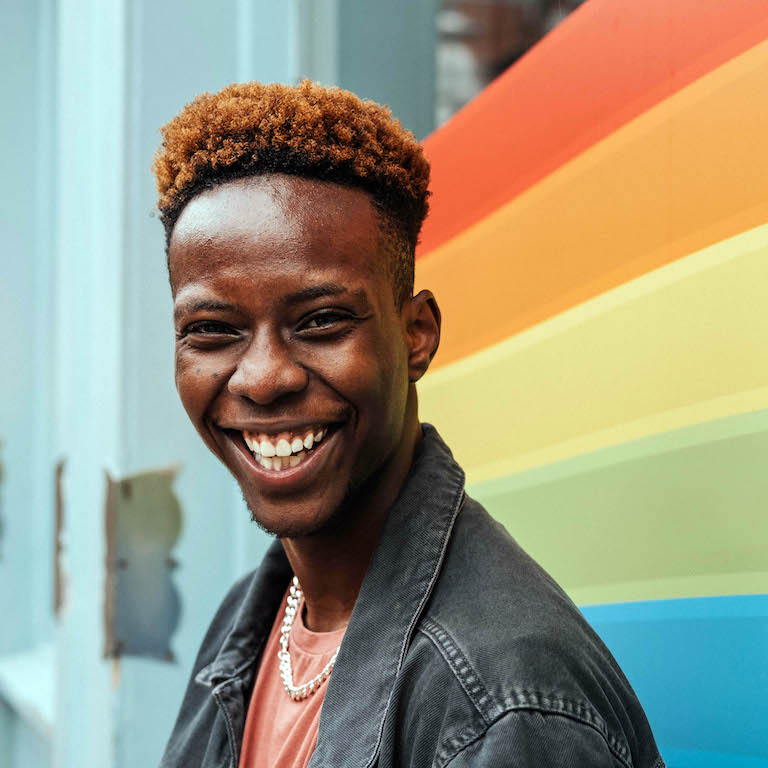} & $\rightarrow$ & \includegraphics[width=2.8cm, height=2.8cm]{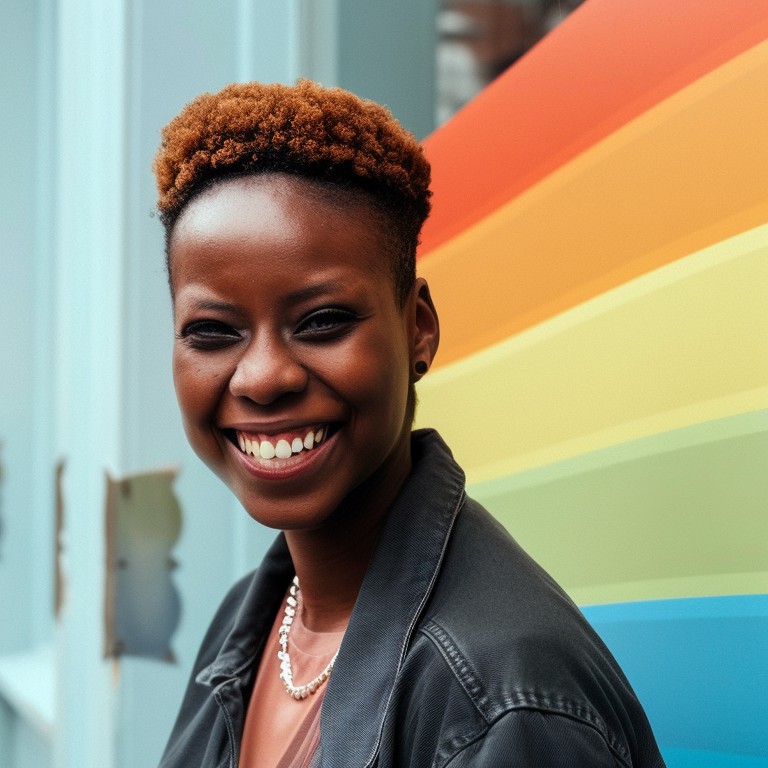} \\
      \multicolumn{3}{c}{``\textit{a \textcolor{BrickRed}{\st{man}} \textcolor{RoyalBlue}{\textbf{woman}} smiling}"} \\
      \includegraphics[width=2.8cm, height=2.8cm]{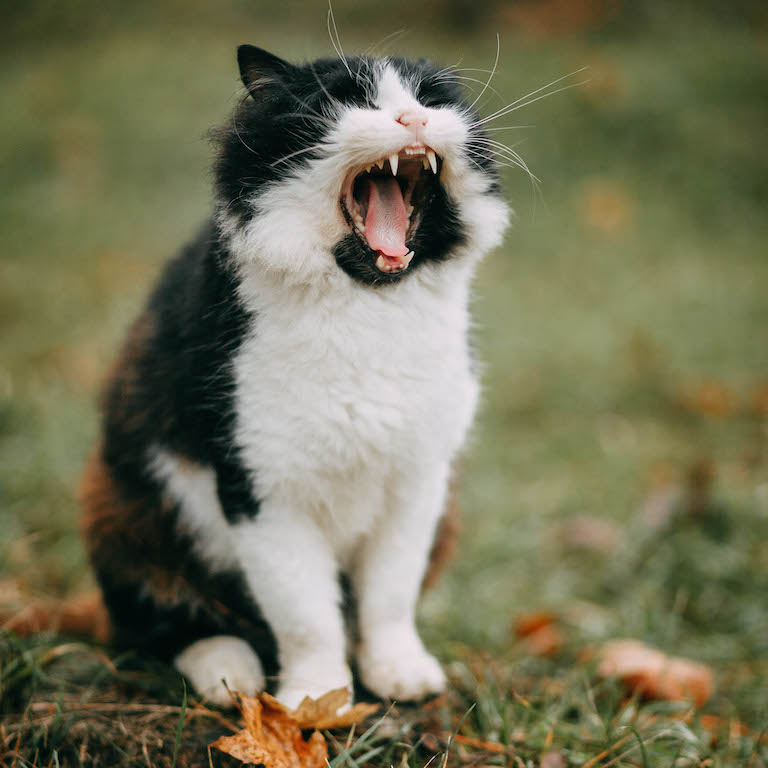} & $\rightarrow$ & \includegraphics[width=2.8cm, height=2.8cm]{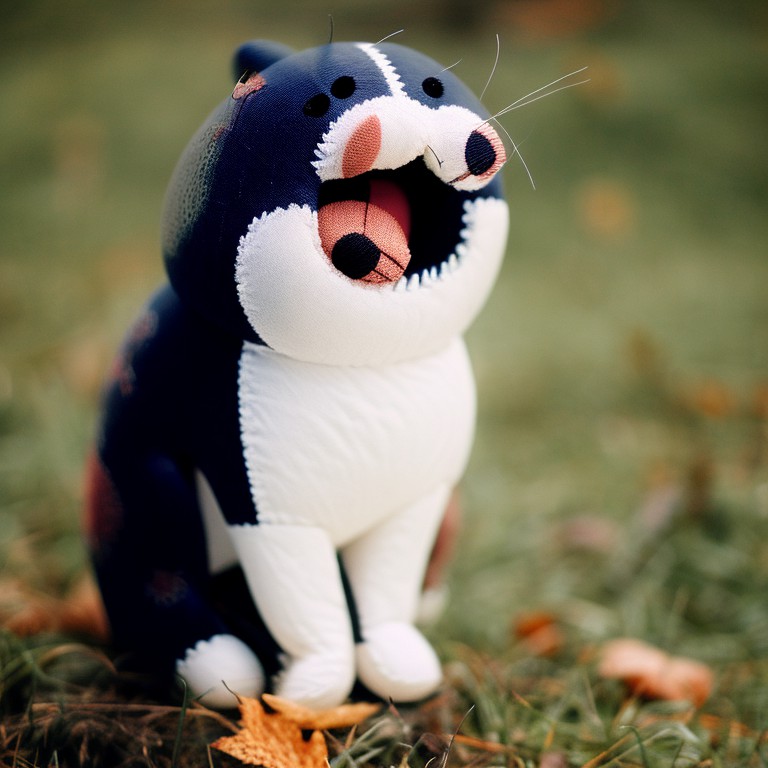} \\
      \multicolumn{3}{c}{``\textit{a \textcolor{RoyalBlue}{\textbf{plush toy}} cat yawning }"} \\
      
    \end{tabular}
    \caption{Prompt-based image editing: the user can add, omit, change, or enhance elements in an image by providing a descriptive prompt of the original image and marking the words that must be removed (in \textcolor{BrickRed}{\textbf{red}}) or added (in \textcolor{RoyalBlue}{\textbf{blue}}).}
    \label{fig:teaser}
  \end{figure}

%% file: sec/2_related.tex
\section{Related Works}
\label{sec:related}

With the impressive advancements in text-to-image diffusion models~\cite{rombach2022latent}, there has been a growing interest in exploring image editing using pre-trained diffusion models. These studies have introduced several editing tasks where the user can guide the generated image through various inputs. For instance, SDEdit~\cite{meng2022sdedit} allows users to apply brush strokes to areas they wish to edit. The model then injects random noise into these targeted areas and uses the diffusion process for denoising. To create new images from examples, techniques like Textual Inversion~\cite{gal2023textual} and Dream-Booth~\cite{ruiz2023dreambooth} employ gradient-descent-based optimization to learn personalized concepts. Text-based editing, in particular, has garnered considerable interest due to its intuitive and user-friendly interaction style. In this domain, DiffusionCLIP~\cite{kim2022diffusionclip} uses DDIM inversion~\cite{song2021ddim} to reverse the diffusion process and applies fine-tuning. This approach guides the generation with a CLIP-based loss to align the generated image more closely with the intended edit.
Another method, as demonstrated in ControlNet~\cite{zhang2023adding}, involves conditioning the generation process on the edges or pose information extracted from the input image. This technique aims to generate an image that retains the original spatial structure yet {\modified{}is styled} according to the given prompt.

  \begin{figure*}[ht]
    \centering
    \includegraphics[width=0.85\linewidth]{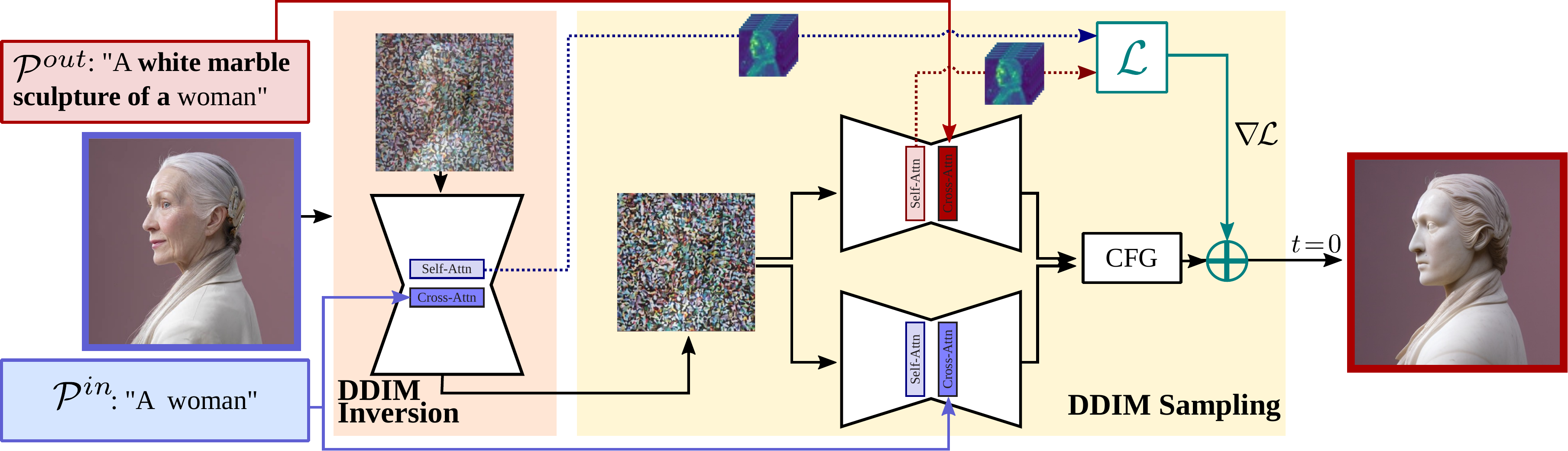}
      
    \caption{
      Pipeline of \method{}: The process begins with DDIM inversion applied to the input image using its corresponding prompt, $\pin$. This inversion yields the estimated noise $\vz_T$, which serves as the initial condition for the DDIM sampling process responsible for generating the edited image. The U-Net processes both the editing prompt $\pout$ and the initial prompt $\pin$ separately, implementing classifier-free guidance. To enhance reconstruction and mitigate inaccuracies introduced by DDIM inversion, a guidance term is computed. This term is derived by comparing the self-attention maps from the DDIM inversion with those estimated by the U-Net when conditioned on the initial prompt $\pin$, ensuring closer alignment and improved fidelity in the final output.
    }
      \label{fig:method_diag}
\end{figure*}
    
{\modified{}This work focuses} on the prompt-based editing task as formulated in \cite{mokady2023null}. In this task (see Fig.~\ref{fig:teaser}) a user provides an image along with a textual prompt  $\pin$, which describes the input image. The user can then instruct the model to add, remove, change, or enhance elements in the image by providing a target prompt $\pout$ corresponding to the desired image (also called positive prompt). This problem formulation has inspired several subsequent studies~\cite{mokady2023null,parmar2023zero,tumanyan2023plug,ju2023direct}.
{\modified{}A high-level comparison is presented}  in Fig.~\ref{fig:related}, where all methods use CFG for conditioned generation and employ a mechanism to reverse the diffusion process, enabling the reconstruction of the input image from Gaussian noise. These approaches vary in their inversion mechanisms and CFG prompting strategies.

In the \nulltextlong{} (\nulltext) (see Fig.~\ref{fig:related}a), Mokady \etal{} \cite{mokady2023null} optimize the null prompt embedding fed to the diffusion model for input reconstruction, with editing facilitated via a cross-attention mechanism \cite{hertz2023prompt}. In \ptopzerolong{} \cite{parmar2023zero} (Fig.~\ref{fig:related}b), the computationally intensive inversion process is bypassed by introducing a guidance term at each diffusion step, steering the model toward accurate reconstruction. Conversely, \negativelong{} (\negative) \cite{miyake2023negative} (Fig.~\ref{fig:related}c) replaces the conventional CFG's null prompt with a negative prompt, with editing achieved through cross-attention manipulation \cite{hertz2023prompt, mokady2023null}. In \directinversionlong{} (\directinversion) \cite{ju2023direct} (Fig.~\ref{fig:related}c), the authors propose a direct inversion method that corrects the reconstruction process and introduces an edit benchmark. 
More recently, EDICT \cite{wallace2023edict} and BDIA \cite{zhang2024exact} have advanced exact diffusion inversion methods using coupled transformations and bidirectional integration approximation, respectively.

%% file: sec/3_method.tex
\section{Method}
\label{sec:method}

{\modified{}This work addresses} the prompt-based image editing task as introduced in \cite{hertz2023prompt,mokady2023null} (see Fig.~\ref{fig:teaser}): the user provides an input image $\vi$ alongside a textual prompt $\pin$ that describes the input image. The user also provides a target prompt $\pout$ which describes the image to obtain after editing.
To address this task, {\modified{}this paper introduces} \method, a method based on Self-Attention Guidance for image Editing whose main pipeline is depicted in Fig.~\ref{fig:method_diag}. 
{\modified{}The proposed approach} (Fig.~\ref{fig:related}d) diverges from the compared methods in two main aspects: (i) {\modified{}it achieves} effective editing without the explicit reconstruction of the input image; other works that do not enforce reconstruction usually achieve good editing performance, but the original image content is poorly preserved \cite{tumanyan2023plug} or {\modified{}requires} additional clues for reasonable structure conservation \cite{xu2023inversion}. (ii) {\modified{}it leverages} intermediate self-attention latent maps computed during the inverse DDIM process to guide the generation, this semantically rich and stable latent space (unlike noisy and arbitrary VAE latent space) enables editing while preserving the content in regions unaffected by the edit. This results in a simpler, more powerful, and computationally efficient method.

{\modified{}The proposed method assumes} a pre-trained text-to-image diffusion model~\cite{ho2020ddpm,saharia2022imagen}. In particular, {\modified{}it employs} a latent diffusion model which operates in the latent space of a pre-trained autoencoder~\cite{rombach2022latent}.
Diffusion models are generative models that employ a neural network as a noise predictor, $\varepsilon_\theta^t(\vz_t,\mathcal{P})$, tasked with the restoration of gradually noised data points at various time steps denoted by ${t \in[0,T]}$. Here, $\vz_t$ represents the noise-altered version of the initial sample $\vz_0$, expressed as $\vz_t = \sqrt{\alpha_t} \vz_0 + \sqrt{1-\alpha_t} \varepsilon$, where $\varepsilon$ is the added Gaussian noise. The noise level is controlled by the variable $\alpha_t$, which ranges from nearly 1, indicating no noise, to almost 0, denoting complete Gaussian noise, as time progresses from $1$ to $T$. Additionally, $\mathcal{P}$ is an optional conditioning variable which, in {\modified{}this} case, takes the form of a textual prompt.  The network $\varepsilon_\theta^t(\vz_t,\mathcal{P})$ is implemented with a U-Net equipped with both self-attention and cross-attention layers~\cite{vaswani2017attention} which process the conditioning information.

{\modified{}Following} previous works~\cite{preechakul2021diffusion, mokady2023null, miyake2023negative}, {\modified{}this paper adopts a widely used} variant of diffusion models known as DDIM that enables faster sampling \cite{song2021ddim}.

{\modified{}A pre-trained encoder} $Enc(\cdot)$ {\modified{}projects} the input image $\vi$ into the latent space, $\zddim_0=Enc(\vi)$, and apply deterministic DDIM inversion \cite{song2021ddim} to reverse the diffusion process. Given the input prompt $\pin$,  DDIM inversion gives {\modified{}a reversed} sequence of noisy latent variables $\zddim_{t}$, where $t$ increases from $0$ to $T$
 
\begin{align}
        \label{eq:ddim_inversion}
        \zddim_{t+1}=&\frac{\sqrt{{\alpha}_{t+1}}}{\sqrt{{\alpha}_{t}}} (\zddim_{t}-\sqrt{1-{\alpha}_{t}}\varepsilon_{\theta}^{t}(\zddim_{t},\mathcal{P}))\notag\\
        &+\sqrt{1-{\alpha}_{t+1}}\varepsilon_{\theta}^{t}(\zddim_{t},\pin)
      \end{align}

      \begin{figure}[h]
  \centering

   \includegraphics[width=0.8\linewidth]{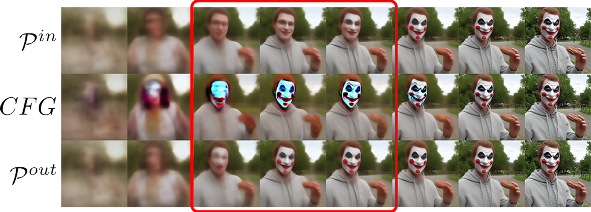}

   \caption{
      Positive and Negative prompt $\hat \vz_0$ estimation across timesteps. We visualize the estimated $\hat \vz_0$ for positive $\pout$, negative prompt $\pin$, and CFG  (both $\pout$ and $\pin$) during DDIM sampling with CFG.
   }
   \label{fig:posneg_direction}
\end{figure}

\subsection{Self-Attention Guidance}
\label{sec:self_attention_guidance}

{\modified{}This paper proposes} two complementary guidance mechanisms to simultaneously achieve effective editing—altering the image to match the target prompt $\pout$—and faithful reconstruction—preserving key regions of the input image.

For editing, {\modified{}a variant of classifier-free guidance (CFG) is adopted} \cite{miyake2023negative}

\begin{align}
  \label{eq:cfg}
    \begin{split}
      \tilde{\varepsilon}_{\theta}^{t}(\vz_{t},&\pin,\pout)= \varepsilon_{\theta}^{t}(\vz_{t},\pin)\\
      &+ w \cdot \left( \varepsilon_{\theta}^{t}(\vz_{t},\pout) - \varepsilon_{\theta}^{t}(\vz_{t},\pin) \right) 
    \end{split}
\end{align}
with $w>0$ modulating the balance between the two objectives. As illustrated in Fig.~\ref{fig:posneg_direction}, early in the diffusion process, the noise estimates for $\pin$ and $\pout$ both steer $\hat{\vz}_t$ toward the original image; then diverge: $\varepsilon_{\theta}^{t}(\vz_{t},\pin)$ favors reconstruction while $\varepsilon_{\theta}^{t}(\vz_{t},\pout)$ drives the editing transformation. 
As $t$ approaches 1, these estimates converge, resulting in a final image that balances input fidelity with the desired prompt edits.
In practice, an excessively low $w$ results in over-reconstruction, whereas a very high $w$ may neglect crucial details of the input image.

To promote reconstruction without resorting to explicit optimization or direct latent-space comparisons (which can cause diffusion instability), {\modified{}the proposed method leverages} the self-attention maps within the U-Net architecture. Unlike cross-attention—which only associates textual tokens with specific image regions \cite{hertz2023prompt,han2023proximal}—self-attention encodes global interactions among all image tokens, capturing richer spatial relationships that are essential for preserving image details not explicitly mentioned in the prompt. Concretely, during DDIM inversion, {\modified{}self-attention maps} $S_{i,t}^{in}$ {\modified{}are recorded} for each transformer block, and during synthesis, {\modified{}corresponding maps} $S_{i,t}^{out}$ {\modified{}are collected} from $\varepsilon_{\theta}^{t}(\vz_{t},\pout)$. {\modified{}Reconstruction is then enforced} by minimizing the loss
\begin{equation}
    \label{eq:self_attention_guidance}
    \mathcal{L}^\text{self}_t =  \sum^N_i \|S_{t,i}^{in} - S_{t,i}^{out}\|_1.
\end{equation}
This loss gradient, scaled by a factor $\lambda$, is incorporated into the noise update
\begin{equation}
    \hat{\vz}_{t-1} = \vz_{t-1} - \lambda \nabla_{\vz_t}\mathcal{L}^\text{self}_t.
\end{equation}
Following \cite{couairon2023monotonical}, {\modified{}factor} $\lambda$ progressively decreases with $t$, so that early diffusion steps emphasize the editing transformation while later steps focus on denoising and fine reconstruction.

In contrast to existing methods that require explicit reconstruction optimization \cite{mokady2023null,miyake2023negative} or depend solely on cross-attention for guidance, {\modified{}the presented approach integrates} self-attention guidance to capture all necessary information from the DDIM inversion. This strategy not only stabilizes the reverse diffusion process but also achieves a more robust and balanced trade-off between editing and reconstruction, thereby differentiating {\modified{}this contribution} from prior work.

\input{fig/crossatt_map_figure.tex}

\subsection{Cross-Attention Manipulation}
\label{sec:cross_attention_manipulation}
Following \cite{hertz2023prompt} (see also \cite{mokady2023null,parmar2023zero,han2023proximal,ju2023direct,miyake2023negative}), {\modified{}this method uses} the U-Net’s cross-attention maps—which link latent space coordinates to prompt tokens—to guide the structural editing process. {\modified{}Proposed} adaptations include three mechanisms:

\textbf{Local Blending.} To recover fine details (e.g., colors and textures) that may be lost with self-attention alone, {\modified{}a high-resolution blending mask is derived} from the cross-attention maps. {\modified{}The maps from both the original prompt} $\pin$ and the editing prompt $\pout$ {\modified{}are aggregated} across diffusion steps and U-Net layers, then normalized, thresholded, and upscaled to obtain a binary mask $M$. This mask is used to fuse the edited latent $\vz_{t-1}$ with the original latent $\zddim_{t-1}$

\begin{equation}
    \label{eq:local_blend}
    \hat{\vz}_{t-1} = M \odot \vz_{t-1} + (1-M) \odot \zddim_{t-1}.
\end{equation}
This approach enhances detail preservation and improves computational efficiency by leveraging a single CFG estimation.

\textbf{Cross-Attention Replacement.} When $\pout$ involves a word swap in $\pin$, {\modified{}the shape of the modified object is preserved} by replacing the cross-attention maps for the altered token. Specifically, for token $k$, {\modified{}the method substitutes} $C_{t,i}^{out}[k]$ with $C_{t,i}^{in}[k]$ during the early diffusion stages (roughly the first 20\%), balancing shape retention with overall image quality.

\textbf{Cross-Attention Reweighting.} Finally, users can adjust the influence of specific words by reweighting the corresponding cross-attention maps $C_{t,i}^{out}[k]$, offering fine-grained control over the editing outcome.

These mechanisms, derived with minor adaptations from \cite{hertz2023prompt}, complement {\modified{}the presented framework} by refining structural guidance and detail preservation during image synthesis.

%% file: fig/crossatt_map_figure.tex
\begin{figure}[h]
    \small
    \centering
    \begin{tabular}{ccc}
      \includegraphics[width=1.8cm, height=1.8cm]{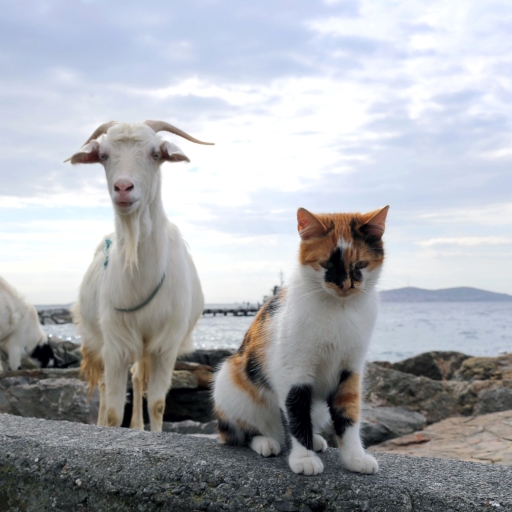} & \includegraphics[width=1.8cm, height=1.8cm]{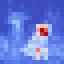} & \includegraphics[width=1.8cm, height=1.8cm]{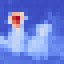} \\
      Input & \textit{``cat"} & \textit{``goat"} \\
      
    \end{tabular}
    \caption{Averaged $16 \times 16$ cross-attention maps corresponding to ``\textit{cat}" and ``\textit{goat}" for the input ``\textit{A cat and a goat}."}
    \label{fig:crossatt_map}
  \end{figure}


%% file: sec/4_experiments.tex
\section{Experiments}
\label{sec:experiments}
Every \method{} experiment was run on a single NVIDIA A100-40GB of a DGX A100 server. All code for generation, experimentation, evaluation, and ablation studies {\modified{}is} available at \codeurl.
The images for remaining methods used in the quantitative comparison (Sec~\ref{sec:expsota}) and user study (Sec~\ref{sec:userstudy}) are taken from the PieBench experimentation files \cite{ju2023direct}. The images for the qualitative comparison (Fig.~\ref{fig:method_comparison} and Sec~\ref{sec:expsota}) corresponding to \nulltext{}, \negative{}, and \negprox{} are taken from \cite{han2023proximal}; the rest were generated with the source code of \cite{ju2023direct}, modifying hyperparameters until the best result was achieved.

As discussed in \cite{mokady2023null,han2023proximal}, all compared methods and diffusion models in general \cite{ho2021classifier} are sensitive to hyperparameters, including {\modified{}\method{}}. Nevertheless, for fairness of comparison, all \method{} results in Tables \ref{tab:ablation_piebench}, \ref{tab:comparison_piebench}, and \ref{tab:user_study} are obtained with the same hyperparameters: 50 DDIM steps, CFG scale of 7.5, local blend in the first 40 steps, cross-attention replacement in the first 5 steps, a 200 self-attention guidance scale, and 2.0 cross-attention reweighting, similar to \directinversion{} \cite{ju2023direct} configuration.

Every method uses $512 \times 512$ images and Stable Diffusion 1.4 as base diffusion model except \pnplong{}, {\modified{}which} uses 1.5. The images in Fig.~\ref{fig:teaser} are $768 \times 768$ images generated using Stable Diffusion 2.1 as the backbone of \method{}.

\input{fig/ablation_figure.tex}

\subsection{Evaluation}

\paragraph{Data} {\modified{}The analysis is conducted} on PieBench \cite{ju2023direct} and MagicBrush \cite{Zhang2023MagicBrush}. PieBench contains 700 images evenly distributed across natural and artificial scenes in four categories (animal, human, indoor, and outdoor) and 10 editing tasks, including object modification, content changes, and style adjustments. Each image includes source and target prompts, edit subjects, and manually annotated masks for background-preservation evaluation (for applicable tasks).  

MagicBrush, designed for instruction-based image editing, consists of 1053 test images with editing instructions; it also supports prompt-to-prompt editing \cite{hertz2023prompt}. Unlike PieBench, it focuses on small-scale edits in photorealistic images.  

Additionally, {\modified{}high-resolution images from Pexels}\footnote{Images in \url{https://www.pexels.com/} are free for commercial use.} {\modified{}are used} for qualitative evaluation.

\paragraph{Metrics} Following {\modified{}the} PieBench evaluation protocol \cite{ju2023direct}, {\modified{}\method{} is evaluated} using three benchmark categories: \textit{1)} structure distance \cite{tumanyan2022splicing}, \textit{2)} background preservation, and \textit{3)} target-prompt fidelity. Background preservation is measured using LPIPS \cite{zhang2018perceptual} and SSIM \cite{wang2004ssim} on the masked area, while prompt fidelity is assessed via CLIP-T Similarity \cite{we2021clipsim} on the entire image and the edited region.  

Similarly, MagicBrush evaluates methods using five metrics: four for input preservation (L1 and L2 distance, CLIP-I similarity \cite{Zhang2023MagicBrush}, and Dino similarity \cite{caron2021emerging}) and one for prompt fidelity (CLIP-T similarity).

\input{fig/quali_comparison.tex}
\subsection{Ablation Study}
\label{sec:expablation}

\input{tables/ablation.tex}

In this ablation study, {\modified{}the evaluation begins} with the mechanism used for achieving reconstruction in the regions that should be preserved through editing. {\modified{}Four baselines are included}, employing either guidance or replacement applied to the cross-attention (CA) or self-attention (SA) layers. The baselines are as follows: (i) guidance based on CA map reconstruction, similar to \cite{parmar2023zero}, and (ii) computing the guidance term in the latent space of the diffusion model, referred to as $\vz_t$ guidance. For SA, {\modified{}two approaches are explored}: (iii) replacement as in \cite{hertz2023prompt,mokady2023null}, and (iv) guidance methods. The proposed method is an enhanced version of (iv), which {\modified{}is further refined} by sequentially integrating: (v) CA replacement, and (vi) Local blending, as detailed in Sec~\ref{sec:cross_attention_manipulation}.

The quantitative results are reported in Table \ref{tab:ablation_piebench}. Due to space constraints and the high correlation among background preservation metrics, {\modified{}the evaluation focuses} on reporting structure distance, LPIPS, and CLIP similarity, specifically in the edited areas. Complementing this quantitative evaluation, {\modified{}the qualitative examples} in Fig.~\ref{fig:ablationquali} {\modified{}showcase} results obtained using the exact same baselines.

Among the various reconstruction mechanisms evaluated, guidance-based approaches (i, ii, and iv) consistently outperform the replacement strategy (iii) across all metrics. This superiority is also reflected qualitatively in Fig.~\ref{fig:ablationquali}, where images resulting from SA replacement are notably unrealistic and diverge from the original. While CA guidance (i) yields satisfactory results, it falls behind {\modified{}the proposed SA guidance} approach by 16 points in the LPIPS metric. Qualitatively, this drop in reconstruction is clearly noticeable in the first row of Fig.~\ref{fig:ablationquali}. Finally, $\vz_t$ guidance (ii) lags behind both in quantitative and qualitative terms compared to the other guidance methods.

{\modified{}Regarding CA replacement (v), observations reveal} that although there is a slight increase in LPIPS, it is compensated by improved structure preservation metrics. This qualitative enhancement is particularly evident in the first row of Fig.~\ref{fig:ablationquali}, where the background preservation is notably better. In (vi), Local Blending (LB) improves both the structure metric and LPIPS, without compromising the CLIP metric. This effectively demonstrates the capability of {\modified{}the LB mechanism} to maintain editability while preserving structural integrity, as it only affects areas unrelated to the editing.

\subsection{Comparison with the State-of-the-Art}

\input{tables/comp_piebench.tex}

\input{tables/comp_magicbrush.tex}

\label{sec:expsota}
\paragraph{Quantitative Comparison}

{\modified{}The proposed method is compared} with state-of-the-art approaches in Tables~\ref{tab:comparison_piebench} and \ref{tab:comparison_magicbrush}. On PieBench, while \ptopzerolong{} underperforms across all metrics, methods such as \pnplong{} obtain high CLIP-T similarity by sacrificing structure and background fidelity, and although \negproxlong{} achieves the best structure fidelity, it does so at the expense of lower CLIP-T similarity. In contrast, \method{} strikes an excellent balance by delivering the best background preservation, near-top (second-best) structure fidelity, the highest Whole CLIP-T, and competitive Edited CLIP-T scores. Similarly, on MagicBrush, \method{} outperforms all alternatives across nearly every metric. These results highlight that \method{} consistently delivers robust, well-rounded performance without relying on inversion of the input image.

\paragraph{Qualitative Comparison}
\label{par:qualitative}
\input{fig/qual_adv.tex}

{\modified{}The qualitative analysis in} Fig.~\ref{fig:method_comparison} is consistent with the quantitative evaluation. {\modified{}The proposed method} preserves the original image structure and content while still achieving good editing performance. For example, only \method{} and \negprox{} are able to preserve the appearance of the tree in the second row. Similarly, \method{} is the only method capable of preserving the details and colors of the t-shirt sleeve, hair, and background in the fourth row.

{\modified{}The proposed method also generates} more natural-looking images. This is especially noticeable in the dog and sushi examples. In the dog example, not only are the dog and the chair preserved, but also the dog's face and the illumination are more natural. In the sushi example, {\modified{}it} is the only method able to preserve all image details and color while generating a deeper red meat color as well as natural-looking \textit{nori} algae, whereas the rest generate shapeless and unnatural-looking sushi pieces. {\modified{}This result is attributed} to the fact that {\modified{}the presented} method is able to diminish the guidance term across time, as discussed in Sec~\ref{sec:self_attention_guidance}.

Furthermore, Fig.~\ref{fig:qual_adv} presents additional examples that showcase the unique strengths of \method{}, which are absent in the compared methods. Unlike these methods, {\modified{}the proposed approach} does not directly guide reconstruction in the $\vz$ latent space (additional experimentation in this regard can be found in supplementary materials' Sec. \ref{sec:sup_reconstruction}). Instead, it operates in the more abstract and semantically rich latent space of the attention layers. {\modified{}It is understood} that as the self-attention maps mostly encode low-resolution semantic features instead of shapes, SAGE is able to generate images that better capture the style while changing the content, has more freedom to edit the shape of elements (as the tulips in the fourth row), and is able to eliminate elements that are present in the original image while filling the gap with plausible content. Moreover, {\modified{}the proposed method} surpasses the compared approaches in transferring style from one image to another while simultaneously altering the content. Conversely, it can preserve content while significantly modifying the style. Notably, all of this is achieved without explicit reconstruction, marking a key distinction from existing methods.

\subsection{User Study}
\label{sec:userstudy}

To strengthen the comparison with existing approaches detailed in Sec.~\ref{sec:expsota}, {\modified{}a user study was conducted}. This study aimed to evaluate {\modified{}the proposed method} against others based on three key aspects: structure preservation, background preservation, and adherence to the prompt. Additionally, {\modified{}overall user preference was assessed}.

{\modified{}A total of 22 participants were recruited} to perform \textit{one-versus-one} comparisons between two randomly selected images from the PieBench \textit{random editing} task. The methods compared included \negativelong~\cite{miyake2023negative}, \directinversionlong~\cite{miyake2023negative}, \negproxlong~\cite{han2023proximal}, \ptopzerolong~\cite{parmar2023zero}, and {\modified{}the proposed method}. The results for most methods showed statistical significance without requiring further experimentation.

However, the comparison between \directinversion{} and \method{} resulted in a particularly tight margin. To ensure statistical significance in this case, an additional 25 participants were later asked to compare these two methods on the MagicBrush dataset, {\modified{}confirming} significance with a \textit{p-value} $< 0.01$ on all methods.

In the study, participants were provided with different sets of images depending on the evaluation criteria. For assessing structure preservation, they were shown the original image, together with two edited versions. In contrast, for background preservation evaluations, the input image was masked to highlight areas relevant to the task. When evaluating prompt fidelity and overall user preference, only the target prompt alongside the edited versions was displayed. Images were presented in a randomized sequence to ensure that participants focused on the relevant aspects for each criterion, and their judgments were unbiased, as they were not informed about which methods were used.

\input{tables/user_study.tex}

The results of these evaluations are compiled in Table~\ref{tab:user_study}. They corroborate {\modified{}the quantitative evaluations}, demonstrating a consistent preference for {\modified{}the proposed method} across all four evaluation criteria. While the preference margin for {\modified{}the method} is somewhat narrow when compared to Direct Inversion in aspects like prompt fidelity and overall preference, a significant difference is observed in terms of background preservation. This pronounced advantage in preserving the background further highlights the effectiveness of {\modified{}the presented approach}. Overall, the user study solidifies the robust performance of {\modified{}the method}, highlighted by the global preference, supporting its strengths not only in quantitative metrics but also in subjective user assessments.

%% file: fig/ablation_figure.tex
 \setlength{\tabcolsep}{1pt}

  \begin{figure*}[h]
    \footnotesize
    \small
    \centering
    \begin{tabular}{ccccccc}
      \hline

      \multirow{2}{*}{Input} &  (i) & (ii) & (iii) & (iv) & (v) & (vi) \\
      &  CA guidance &  $\vz_t$  guidance & SA replace.&  SA guidance &   (iv)+CA replace. &  (v) + LB \\
      \hline
      \multicolumn{7}{c}{``\textit{a girl with blonde hair \textcolor{RoyalBlue}{\textbf{and a fluffy rainbow hat}} smiling}"} \\
      \includegraphics[width=2.1cm, height=2.1cm]{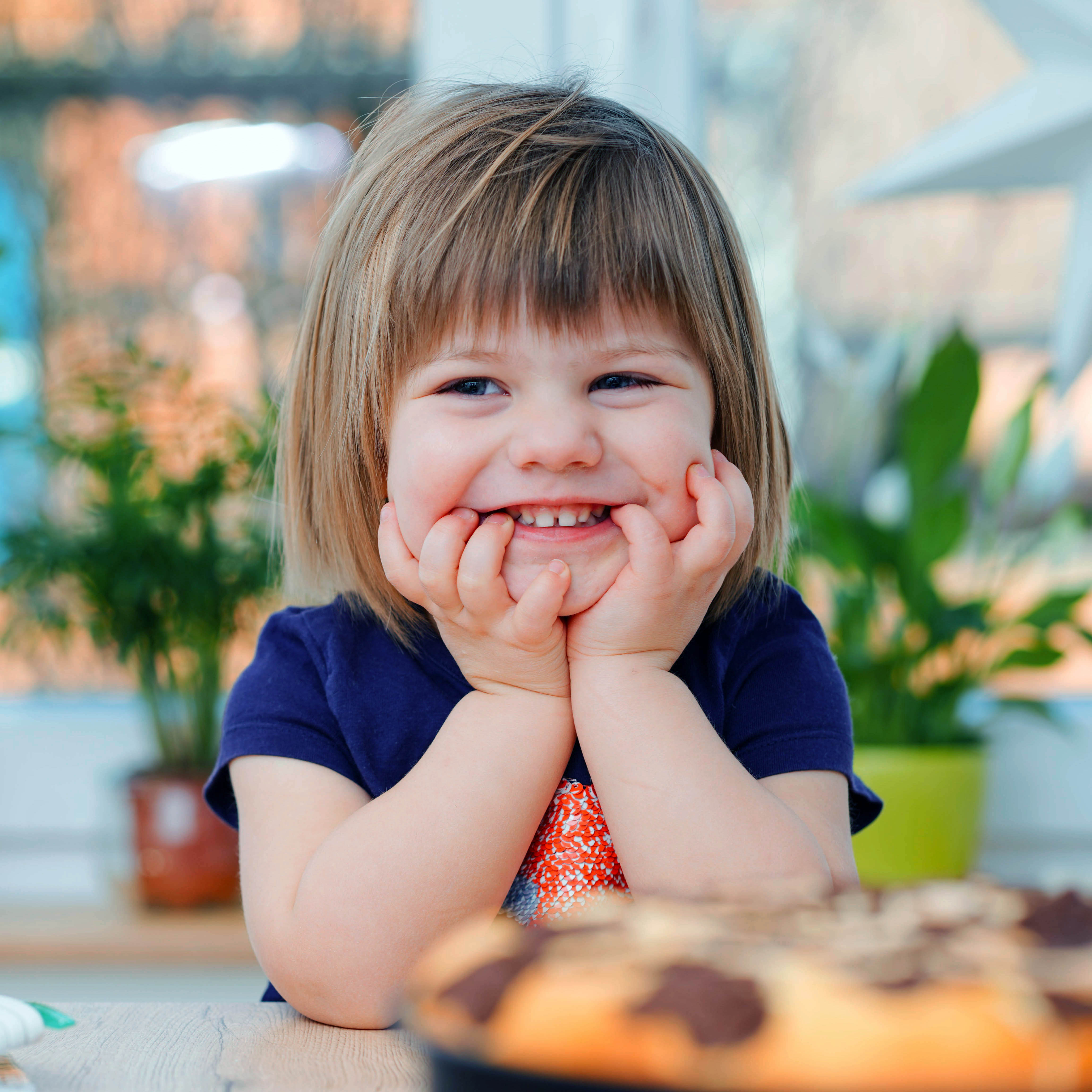} & \includegraphics[width=2.1cm, height=2.1cm]{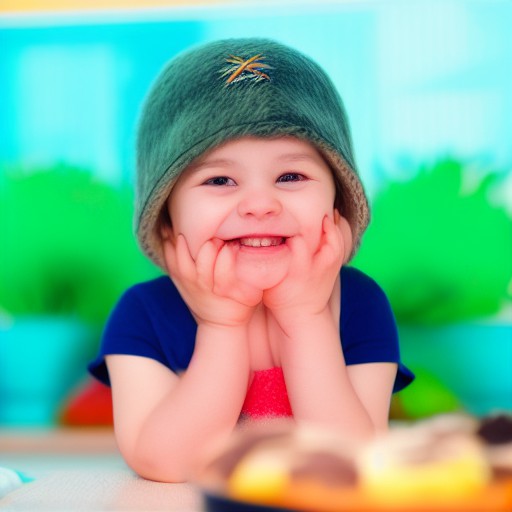} &\includegraphics[width=2.1cm, height=2.1cm]{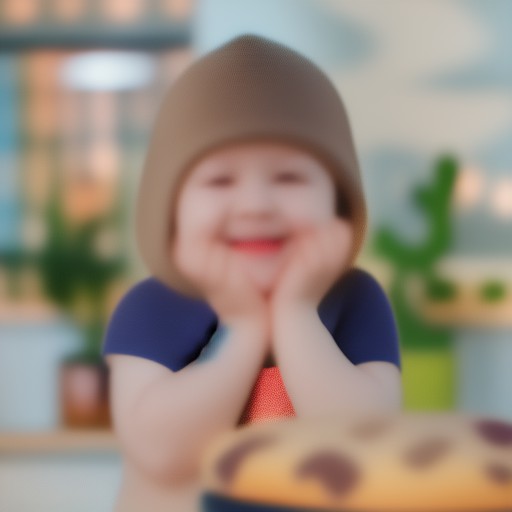} &\includegraphics[width=2.1cm, height=2.1cm]{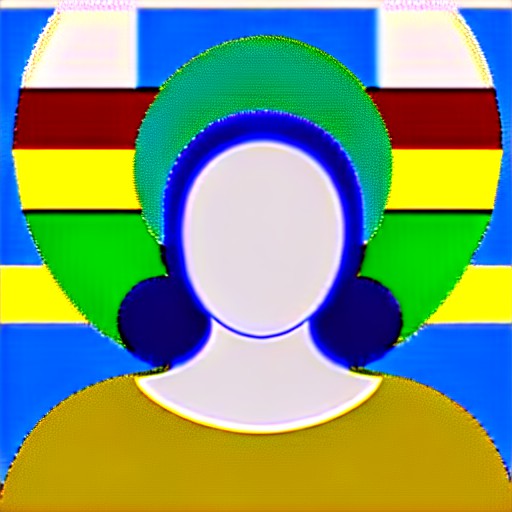} &\includegraphics[width=2.1cm, height=2.1cm]{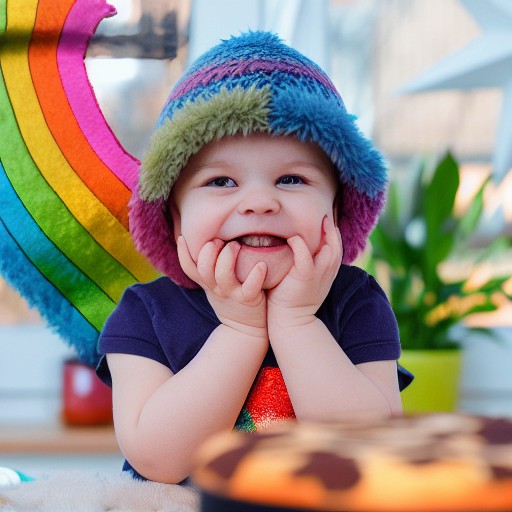} &\includegraphics[width=2.1cm, height=2.1cm]{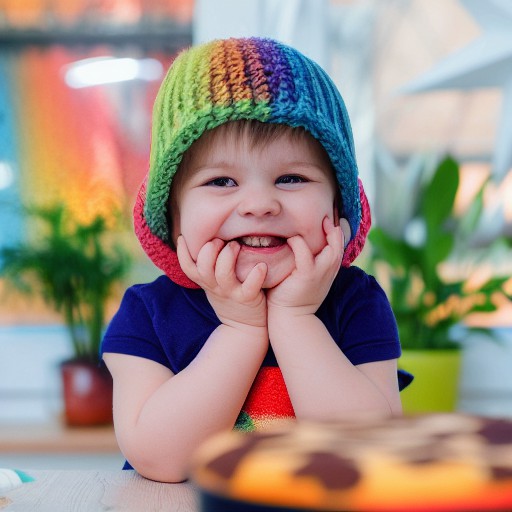} &\includegraphics[width=2.1cm, height=2.1cm]{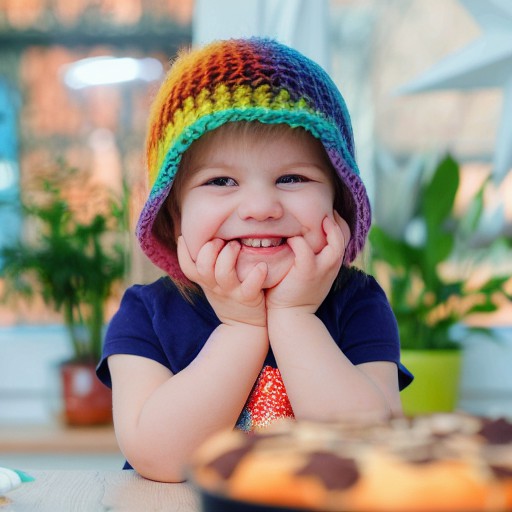} \\
      \multicolumn{7}{c}{``\textit{a baby with a stuffed \textcolor{BrickRed}{\st{monkey}} \textcolor{RoyalBlue}{\textbf{zebra}} in a car}"} \\
      \includegraphics[width=2.1cm, height=2.1cm]{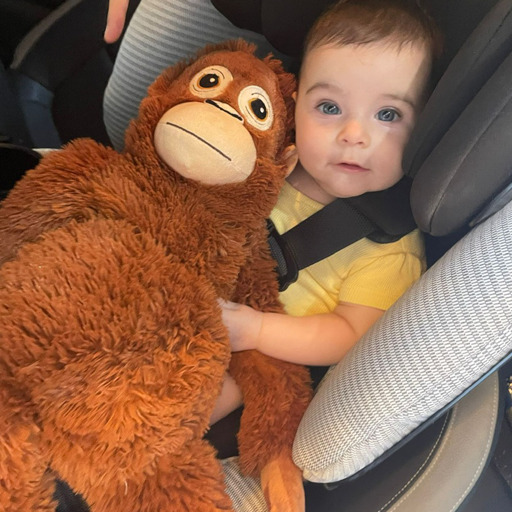} & \includegraphics[width=2.1cm, height=2.1cm]{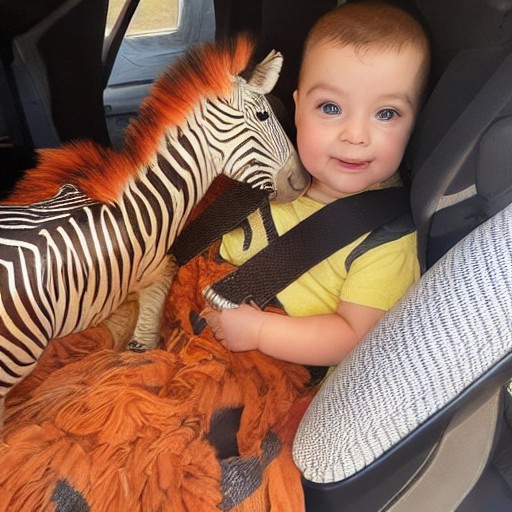} &\includegraphics[width=2.1cm, height=2.1cm]{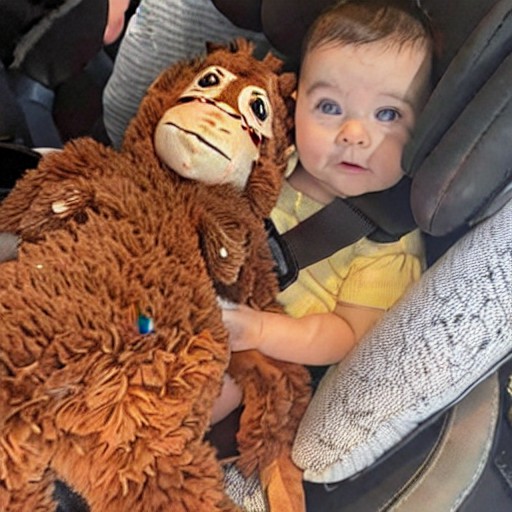} &\includegraphics[width=2.1cm, height=2.1cm]{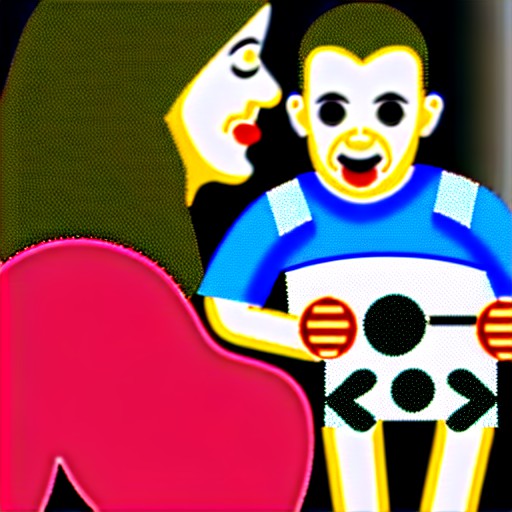}
      &\includegraphics[width=2.1cm, height=2.1cm]{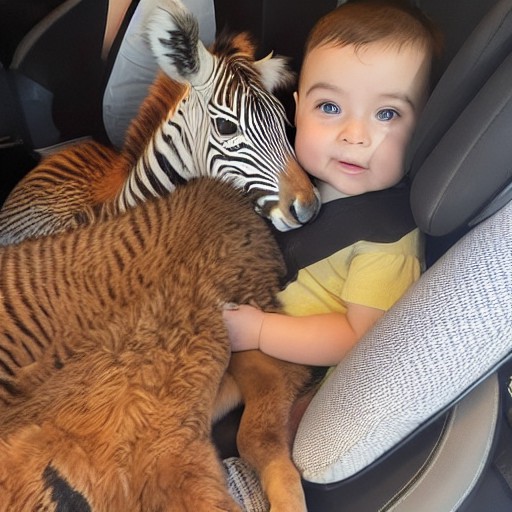} &\includegraphics[width=2.1cm, height=2.1cm]{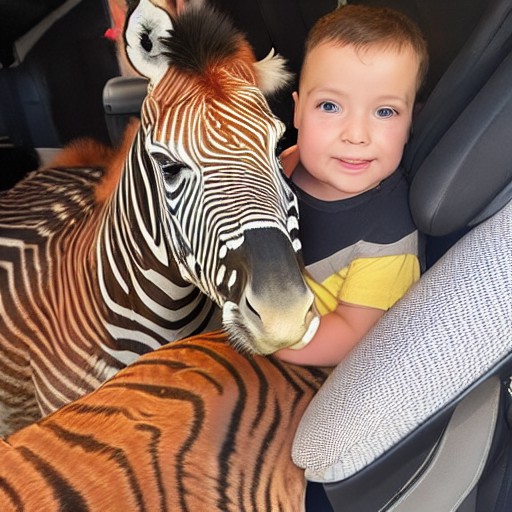} &\includegraphics[width=2.1cm, height=2.1cm]{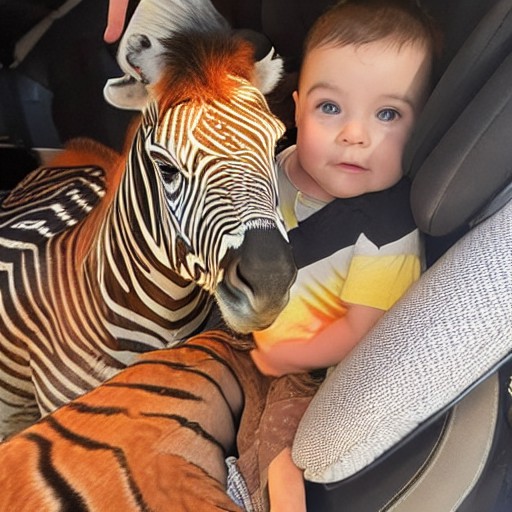} \\
      \hline
    \end{tabular}
    \caption{{\modified{}Qualitative results of the} ablation study. Each column corresponds to a configuration described in Table \ref{tab:ablation_piebench}.}
    \label{fig:ablationquali}
  \end{figure*}

%% file: fig/quali_comparison.tex
\setlength{\tabcolsep}{1pt}
\begin{figure*}[h]
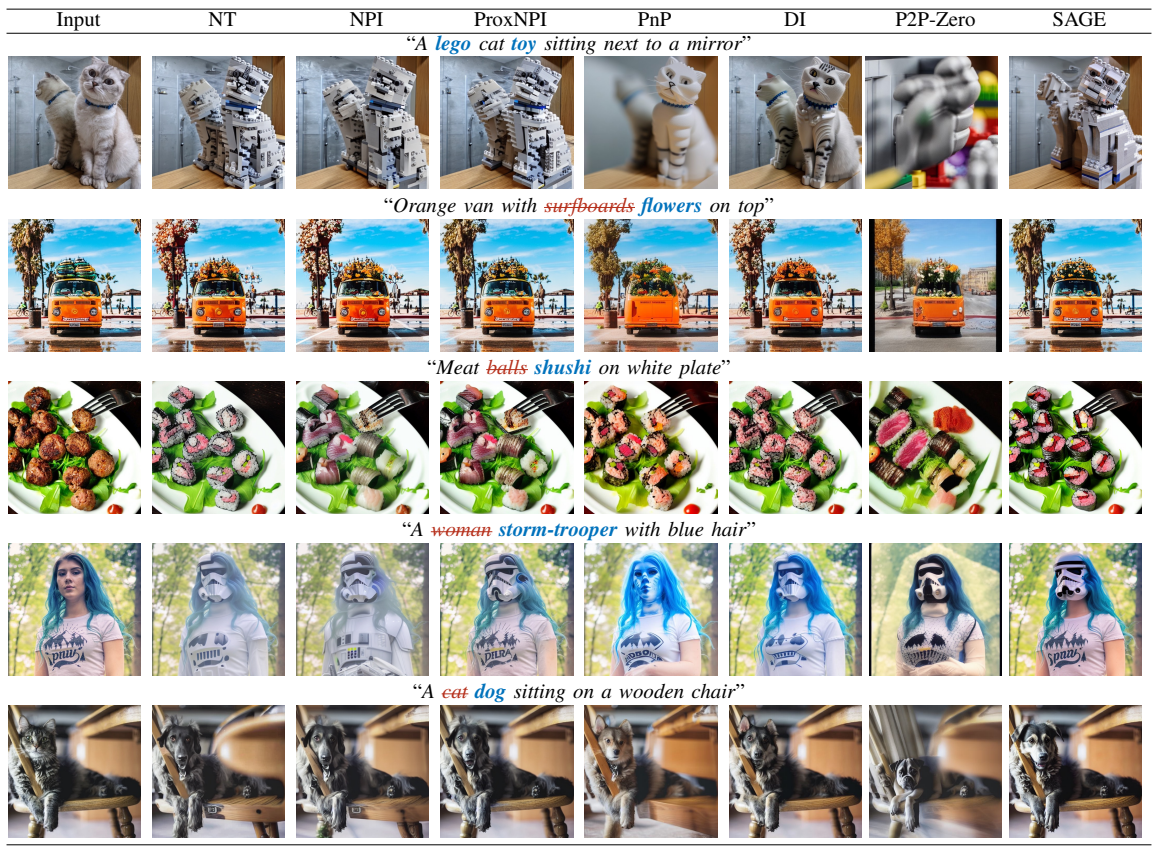

  \centering
    \footnotesize
    \small
    \resizebox{0.85\textwidth}{!}{
    \begin{tabular}{ccccccccc}
      \hline
      Input & \nulltext & NPI & \negprox & PnP & \directinversion & \ptopzero & \method \\
      \hline
      \multicolumn{8}{c}{``\textit{A \textcolor{RoyalBlue}{\textbf{lego}} cat \textcolor{RoyalBlue}{\textbf{toy}} sitting next to a mirror}"} \\
      \insertimage{cat}{input} & \insertimage{cat}{null} &\insertimage{cat}{neg} &\insertimage{cat}{prox} &  \insertimage{cat}{pnp} & \insertimage{cat}{direct}& \insertimage{cat}{zero} & \insertimage{cat}{ours_1}  \\
      \multicolumn{8}{c}{``\textit{Orange van with \textcolor{BrickRed}{\st{surfboards}} \textcolor{RoyalBlue}{\textbf{flowers}} on top}"} \\
      \insertimage{van}{input} & \insertimage{van}{null} &\insertimage{van}{neg} &\insertimage{van}{prox} & \insertimage{van}{pnp} & \insertimage{van}{direct}& \insertimage{van}{zero}& \insertimage{van}{ours}  \\
      \multicolumn{8}{c}{``\textit{Meat \textcolor{BrickRed}{\st{balls}} \textcolor{RoyalBlue}{\textbf{shushi}} on white plate}"} \\
      \insertimage{meat}{input} & \insertimage{meat}{null} &\insertimage{meat}{neg} &\insertimage{meat}{prox} & \insertimage{meat}{pnp} & \insertimage{meat}{direct}& \insertimage{meat}{zero}& \insertimage{meat}{ours}  \\
      \multicolumn{8}{c}{``\textit{A \textcolor{BrickRed}{\st{woman}} \textcolor{RoyalBlue}{\textbf{storm-trooper}} with blue hair}"} \\
      \insertimage{girl}{input} & \insertimage{girl}{null} &\insertimage{girl}{neg} &\insertimage{girl}{prox} & \insertimage{girl}{pnp} & \insertimage{girl}{direct}& \insertimage{girl}{zero3}& \insertimage{girl}{ours}  \\
      \multicolumn{8}{c}{``\textit{A \textcolor{BrickRed}{\st{cat}} \textcolor{RoyalBlue}{\textbf{dog}} sitting on a wooden chair}"} \\
      \insertimage{cat2}{input} & \insertimage{cat2}{null} &\insertimage{cat2}{neg} &\insertimage{cat2}{prox} & \insertimage{cat2}{pnp} & \insertimage{cat2}{direct}& \insertimage{cat2}{zero}& \insertimage{cat2}{ours}  \\
      \hline
    \end{tabular}
    }
    \caption{Qualitative comparison with state-of-the-art {\modified{}methods}. {\modified{}Examples are shown for} both word insertion and word swap.}

    \label{fig:method_comparison}
  \end{figure*}

%% file: tables/ablation.tex
\begin{table}[htbp]
    \centering
    \small
      \renewcommand\arraystretch{0.8}
      \setlength{\tabcolsep}{0.5mm}{
    \resizebox{\linewidth}{!}{
    \begin{tabular}{llcccccc}
     \toprule
          &\multirow{2}{*}{\textbf{Reconstruction}}&\textbf{CA}&\multirow{2}{*}{\textbf{LB}}&\multirow{2}{*}{ \textbf{Struct.}$\downarrow$}      &  \multirow{2}{*}{\textbf{LPIPS} $\downarrow$}     & \multirow{2}{*}{\textbf{CLIP} $\uparrow$} \\ 
 & & \textbf{replace.}&&&&\\
    \midrule
    
    (i)&CA guidance &- & -& 15.7 & 58 & 21.9 \\
    (ii)&$\vz_t$ guidance &-& -& 40.0 & 111.3 & 21.5 \\
    (iii)&SA replace &- & -&178.4 & 376.0 & 18.8 \\
    (iv)& SA guidance& -& -&  15.7 & 42.0 & \textbf{22.0}  \\
    \midrule
    (v)& SA guidance&\checkmark&- & 14.7 & 49.5 & 21.9 \\
    (vi)& SA guidance&\checkmark&\checkmark & \textbf{11.0} & \textbf{39.6} & \textbf{22.0}\\
    \bottomrule
    \end{tabular}}
  
    }
        \caption{Quantitative analysis performed on PieBench \cite{ju2023direct}. {\modified{}Four strategies (i–iv) are evaluated} for guiding reconstruction, based on either guidance or replacement, applied to the cross-attention (CA) or self-attention (SA) layers. {\modified{}Additionally, (v) CA replacement and (vi) Local Blending (LB) are evaluated} in combination with the best reconstruction approach.}
            \label{tab:ablation_piebench}
    \end{table}

%% file: tables/comp_piebench.tex
    \begin{table}[htbp]
    \centering
    \small
      \renewcommand\arraystretch{0.9}
      \setlength{\tabcolsep}{0.5mm}{
  \resizebox{\linewidth}{!}{  \begin{tabular}{cccccc}
    \toprule
          & \textbf{Struct.}          & \multicolumn{2}{c}{\textbf{BG}} & \multicolumn{2}{c}{\textbf{CLIP-T Similariy}} \\ \midrule
    \textbf{Method}            & \textbf{Dist.} $\downarrow$ &   \textbf{LPIPS} $\downarrow$  & \textbf{SSIM} $\uparrow$    & \textbf{Whole}  $\uparrow$          & \textbf{Edited}  $\uparrow$       \\ \midrule
    \textbf{\nulltextlong} \cite{mokady2023null} & 13.4             & 60.7    & 84.1 & 24.8        & 21.9          \\
    \textbf{Negative prompt} \cite{miyake2023negative} &     16.2        &  69.0   & 83.4 &   24.6     &    21.9     \\
    \textbf{\negproxlong} \cite{miyake2023negative}  & \textbf{7.4}  &	42.0  &	\textbf{86.0} &	24.3 &	21.4 \\
    \textbf{\pnplong} \cite{tumanyan2023plug}  & 28.2   & 113.5 & 79.0 & 25.4 & \textbf{22.6} \\
    \textbf{\directinversionlong} \cite{ju2023direct} & 11.7             & 54.6    & 84.8 & 25.0         & 22.1          \\
    \textbf{\ptopzerolong} \cite{parmar2023zero} & 61.7   & 172.2 & 74.7 & 22.8 & 20.5 \\
    \textbf{\method} (ours) & 11.0 &	\textbf{39.6} &	\textbf{86.0} &	\textbf{25.5} &	22.0 \\
    
    \bottomrule
    \end{tabular}}

    }
        \caption{Quantitative analysis performed on PieBench. All rows but ours are taken directly from the \directinversion{} work \cite{ju2023direct}. BG stands for Background.}
            \label{tab:comparison_piebench}
    \end{table}

%% file: tables/comp_magicbrush.tex
    \begin{table}[htbp]
        
    \centering
    \small
      \renewcommand\arraystretch{0.9}
      \setlength{\tabcolsep}{0.5mm}{
  \resizebox{\linewidth}{!}{  \begin{tabular}{cccccc}
    \toprule
    \textbf{Method}            & \textbf{L1} $\downarrow$ &   \textbf{L2} $\downarrow$  & \textbf{CLIP-I} $\uparrow$    & \textbf{Dino}  $\uparrow$          & \textbf{CLIP-T}  $\uparrow$       \\ \midrule
    \textbf{\negproxlong} \cite{miyake2023negative}  & 7.0  &	1.85  &	88.7 &	83.0 &	26.7 \\
    \textbf{\directinversionlong} \cite{ju2023direct} & 8.1             & 2.04    & 89.8 & 84.9         & \textbf{27.9}          \\
    \textbf{\ptopzerolong} \cite{parmar2023zero} & 17.4   & 7.6 & 80.7 & 69.3 & 26.4 \\
    \textbf{\method} (ours) & \textbf{6.4} &	\textbf{1.8} &	\textbf{90.9} &	\textbf{85.9} &	27.6 \\
    
    \bottomrule
    \end{tabular}}

    }
    \caption{    
    Quantitative analysis performed on MagicBrush. 
        }
            \label{tab:comparison_magicbrush}
    \end{table}

%% file: fig/qual_adv.tex
\setlength{\tabcolsep}{1pt}
\begin{figure}[h]
    \footnotesize
    \small
    \begin{tabularx}{\columnwidth}{ YYYYYY }
        Input & \ptopzero &  \nulltext &  \directinversion &  \negprox &  \method \\
        \multicolumn{6}{c}{\includegraphics[width=\columnwidth]{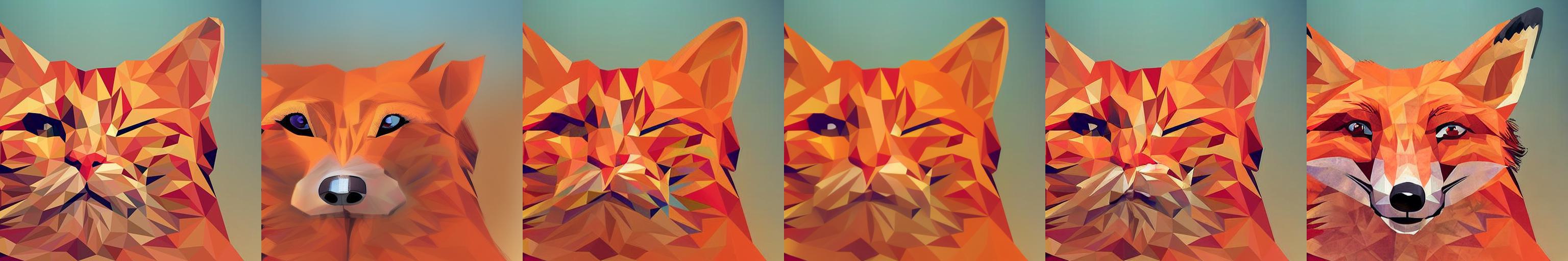}} \\
        \multicolumn{6}{c}{``\textit{a \textcolor{BrickRed}{\st{cat}} \textcolor{RoyalBlue}{\textbf{fox}} is shown in a low polygonal style}"} \\
        \multicolumn{6}{c}{\includegraphics[width=\columnwidth]{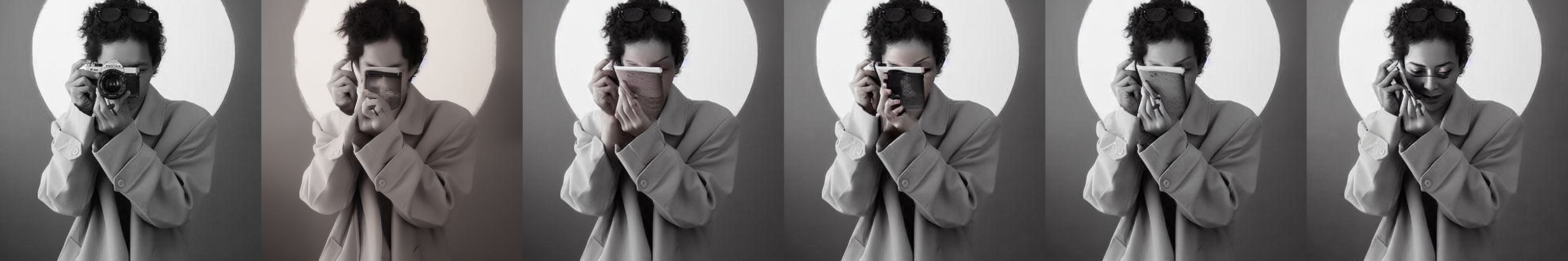}} \\
        \multicolumn{6}{c}{``\textit{a woman in a coat holding a \textcolor{BrickRed}{\st{camera}} \textcolor{RoyalBlue}{\textbf{phone}} on white plate}"} \\
        \multicolumn{6}{c}{\includegraphics[width=\columnwidth]{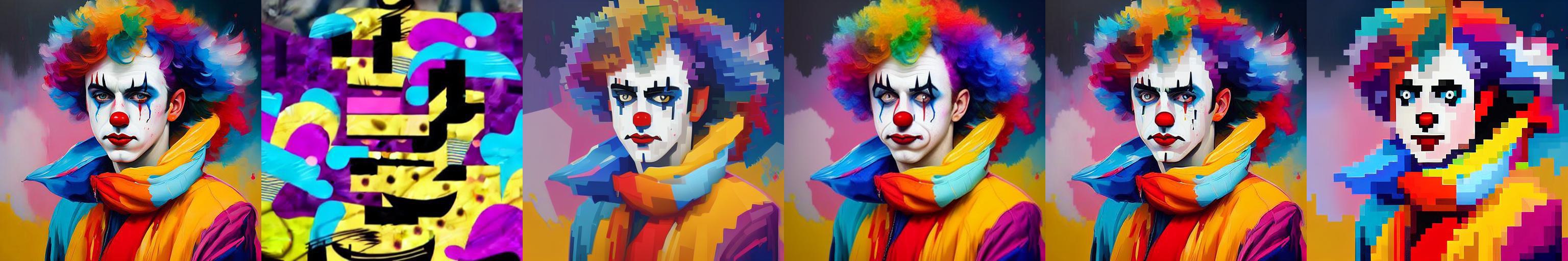}} \\
        \multicolumn{6}{c}{``\textit{a clown \textcolor{RoyalBlue}{\textbf{in pixel art style}} with colorful hair}"} \\
        \multicolumn{6}{c}{\includegraphics[width=\columnwidth]{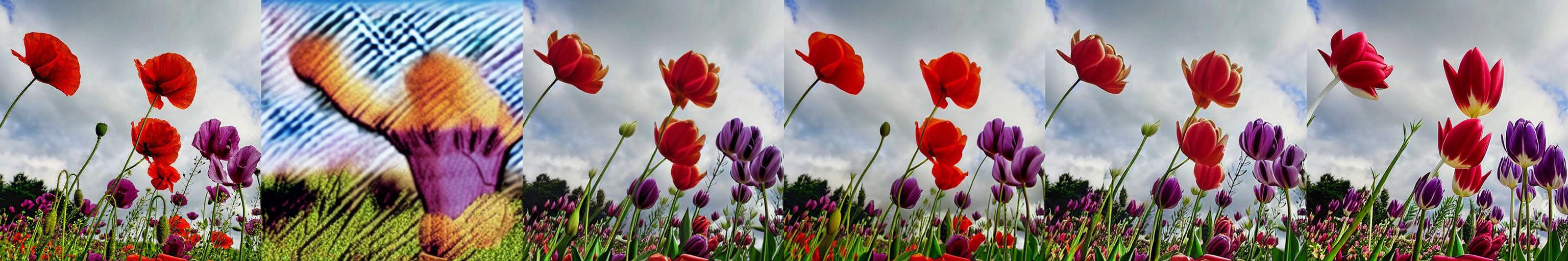}} \\
        \multicolumn{6}{c}{``\textit{\textcolor{BrickRed}{\st{poppies}} \textcolor{RoyalBlue}{\textbf{tulips}}}"} \\
        \multicolumn{6}{c}{\includegraphics[width=\columnwidth]{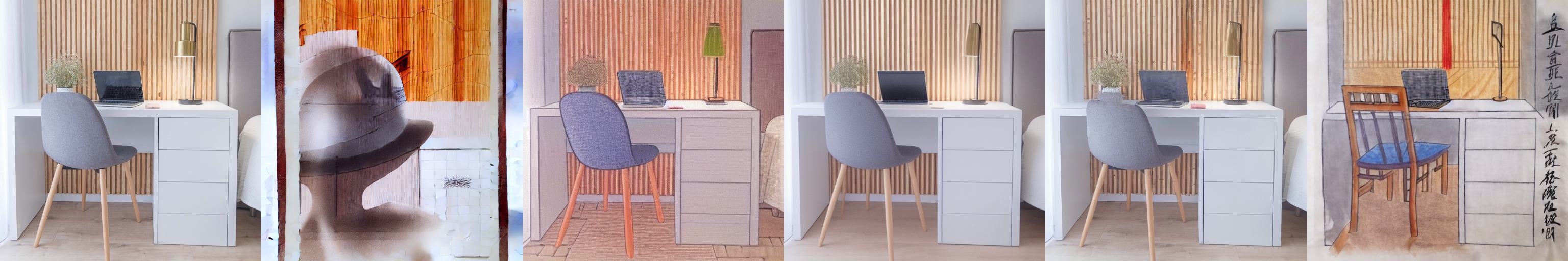}} \\
        \multicolumn{6}{c}{``\textit{\textcolor{RoyalBlue}{\textbf{chinese painting of}} a white desk with a laptop and chair}"} \\
        \end{tabularx}
        \caption{Examples from the PieBench dataset illustrating the strengths of \method{}, including superior style transfer, content modification, and object removal.        }
    \label{fig:qual_adv}
  \end{figure}

%% file: tables/user_study.tex
{

\begin{table}[htbp]
    
    \small
    \centering
      \renewcommand\arraystretch{1.}
      \setlength{\tabcolsep}{0.7mm}{
    \begin{threeparttable}
    \begin{tabular}{ccccc}

    \toprule
    \textbf{\method~vs}& \textbf{Structure\%}  &  \textbf{Background\%} & \textbf{Prompt\%} & \textbf{Global\%} \\
    \midrule
    \ptopzero~\cite{parmar2023zero} &       93.8 &	            92.0 &	    83.0 &	    75.9 \\
    \nulltext~\cite{mokady2023null} &       70.5 &	            69.6 &	    52.7 &	    54.5  \\
     \directinversion~\cite{ju2023direct} & 55.4 / 52.0 &	    62.5 / 55.3 &	    52.7 / 46.5 &	    52.7 / 52.7 \\
         \negprox~\cite{han2023proximal} &  58.9 &	            58.0 &	    62.5 &	    59.8  \\
    \bottomrule
    \end{tabular}
    \end{threeparttable}
    }

\caption{User study results showing the frequency with which {\modified{} the proposed method (\method{})} was preferred over {\modified{} compared methods}. A total of 1792 questions were answered by participants. For \directinversion{}, two results are reported: the first corresponds to the PieBench dataset{\modified{}}, consistent with other methods, while the second represents the user study conducted on MagicBrush with 1540 additional questions to assess statistical significance. All method comparisons demonstrate statistical significance with $p < 0.01$, determined using a binomial test and the \textit{Fisher} method for {\modified{}combining} \textit{p-values}.}

            \label{tab:user_study}
    \end{table}

    }

%% file: sec/5_limitations.tex
\subsection{Limitations}
\label{sec:limitations}

The main limitation of \method{}, as {\modified{}with} other diffusion-based editing methods \cite{han2023proximal}, is its hyperparameter sensitivity. Adjusting hyperparameters for each image enhances quality but affects user experience, making it hard to standardize a single parameter set for diverse editing tasks. {\modified{}Specific failure cases} of \method{} are discussed in the supplementary materials.

%% file: sec/6_conclusions.tex
\section{Conclusions}

{\modified{}This work revisited} the conventional approach to prompt-based image editing within diffusion models. Contrary to established methods that leverage both negative and positive branches of Classifier-Free Guidance for editing ($\pin$ and $\pout$, respectively), {\modified{}the presented investigation reveals} that reconstruction is not necessary. {\modified{}The DDIM inversion process alone contains} enough information for effective editing, thereby questioning the need for manipulating $\pin$ and $\pout$ attention maps. {\modified{}The proposed approach simplifies} the editing process by applying guidance exclusively to the $\pin$ branch, which not only streamlines the method but also yields better results, as confirmed by extensive comparative analyses.

{\modified{}The primary contribution lies} in introducing and validating self-attention guidance as a superior mechanism for image editing tasks. Through quantitative analyses, ablation studies, and user feedback, {\modified{}it was demonstrated} that self-attention guidance captures a broader contextual understanding, enabling better edits compared to traditional cross-attention techniques. This approach preserves closer fidelity to the original image content while accurately applying the desired edits. 

{\modified{}A comprehensive comparative analysis}, supported by an extensive user study, shows that SAGE is preferred over every other compared method with statistical significance (\textit{p-value} $< 0.01$), achieving an average preference higher than 60.7\% among participants. This substantial margin underscores SAGE's effectiveness and potential to redefine standard practices in image editing with diffusion models.


{\modified{}Several directions for improvement and future research have been identified. For object removal, masking out self-attention guidance could mitigate unintended structure preservation. Investigating alternative sampling schedulers beyond DDIM and extending SAGE to higher-quality models, such as distilled models~\cite{meng2023distillation} or rectified-flow models~\cite{liu2023flow,deng2024fireflow}, may further enhance results while reducing the number of sampling steps required. Additionally, dynamically adapting hyperparameters based on task complexity, input characteristics, and denoising magnitude could improve robustness across a wide range of editing scenarios.}

%% file: sec/7_acknowledgements.tex
\section*{Acknowledgements}
\label{sec:acknowledgements}

This work was supported by the Spanish Ministry of Science, Innovation, and Universities (MICIU) under grant FPU19/00591, grant CONFIA (PID2021-122916NB-I00) funded by MICIU/AEI/10.13039/501100011033 and by “ERDF A way of making Europe” and by the French National Research Agency (ANR- 20-CE23-0027).

%% file: sec/X_suppl.tex
\clearpage
\setcounter{page}{1}
\maketitlesupplementary


The supplementary materials provide additional analysis to enrich the main paper, along with further results and insights. This document is structured as follows: Section~\ref{sec:sup_sourcecode} contains the source code and demo resources, with implementation information provided in Section~\ref{sec:sup_sourcecode_details}. Section~\ref{sec:sup_results} presents additional experimental outcomes including time and memory requirements (Section~\ref{sec:sup_results_time}), guidance ablations (Section~\ref{sub:sup_results_ablation}), and an evaluation of reconstruction performance (Section~\ref{sec:sup_reconstruction}). Finally, Sections~\ref{sec:sup_limitations_structure} and~\ref{sec:sup_limitations_failure} discuss limitations regarding structure preservation and failure cases.

\section{Source Code}
\label{sec:sup_sourcecode}

The source code, scripts for replicating the experiments, and a web demo (including an offline version via the Gradio app and an online version hosted on Hugging Face) are available at \codeurl{}. Additionally, the PIE-Bench and MagicBrush images used for the quantitative results in Tables \ref{tab:ablation_piebench}, \ref{tab:comparison_piebench}, and \ref{tab:user_study} of the main paper are also accessible. These images are provided in PNG format, with all editing parameters embedded as metadata, facilitating further independent analysis.

\subsection{Implementation Details}
\label{sec:sup_sourcecode_details}

In our preliminary analysis, best results for $512 \times 512$ images were achieved with $32 \times 32$ self-attention maps and $16 \times 16$ cross-attention maps, particularly from the second and third encoder blocks of the U-Net respectively and the corresponding upsampling block, similar to \cite{hertz2023prompt}. For $ 768 \times 768 $ images, the best results were obtained with $ 24 \times 24 $ self-attention and cross-attention maps, specifically from the third block. Additionally, our code supports FP16 computation, including gradient calculation, which significantly reduces time and memory consumption. To prevent gradients from becoming zero in half-precision floating-point arithmetic, the loss term is scaled by a factor of 500 prior to gradient computation, with the weighting factor $\lambda$ applied afterward (Sec.~\ref{sec:self_attention_guidance}).

\section{Additional Results}
\label{sec:sup_results}

\subsection{Inference time and memory comparison}
\label{sec:sup_results_time}

\input{tables/comp_time.tex}

Table~\ref{tab:comparison_time} summarizes the performance of various prompt-based editing methods on a NVIDIA A100-40GB GPU. The Time column reports the additional time required to generate a second 512×512 image (isolating generation cost from model/data loading), and the Memory column indicates the peak memory allocated per image as measured using the command-line utility \texttt{nvidia-smi}.

Our FP16 version of \method{} uses the least memory (7.4 GB) and is the second fastest (12.6 s), while the FP32 version also remains competitive. These advantages are primarily due to two factors: (i) gradient-based guidance is computed on only a subset of self-attention maps, greatly reducing computational load, and (ii) the complete omission of the input image reconstruction step streamlines processing and lowers memory usage. All methods were executed under identical conditions using their original codebases.

\input{fig/cfg_self_matrix.tex}

\subsection{Classifier-Free and Self-attention guidance} 

\label{sub:sup_results_ablation}

Figure~\ref{fig:supablation_cfg_guidance} evaluates the impact of both classifier-free guidance ($w$) and self-attention guidance ($\lambda$) on the generation process. It can be seen how a higher $\lambda$ value results in greater attention to the input structure, whereas greater $w$ values result in more plausible generation but also in more saturated colors (already discussed in \cite{ho2021classifier}). Higher self-attention guidance $\lambda$ values result in both better structure preservation and better color preservation as it reduces the negative impact of high CFG in natural looking colors. An appropiate balance is necessary to generate images that preserve the structure of the input image while performing deep, natural-looking transformations that reflect $\pout$.

\input{fig/res_reconstruction}

\input{fig/res_reconstruction_table.tex}

\subsection{Self-attention guidance for reconstruction}
\label{sec:sup_reconstruction}
Figure~\ref{fig:res_reconstruction} compares the reconstruction performance of \method{} given the same input image and editing prompt against the baseline VAE reconstruction. This evaluation applies only self-attention guidance, without local blending or auxiliary techniques. The results clearly show that SAGE preserves fine details despite not explicitly optimizing for reconstruction like \nulltextlong{} or compensating for VAE reconstruction errors as \directinversionlong{} does. Although our method performs worse on strict reconstruction metrics (Table~\ref{tab:res_reconstruction}) due to its indirect approach, this has no noticeable impact on final editing quality. As shown in Tables~\ref{tab:comparison_piebench}, \ref{tab:comparison_magicbrush}, and \ref{tab:user_study}, the proposed method outperforms the compared methods, particularly in reconstruction-related evaluations.

\section{Limitations}
\subsection{Structure Preservation}
\label{sec:sup_limitations_structure}

    Figure~\ref{fig:sup_structpres} shows that when objects are removed by omitting corresponding words from the prompt, \method{} fills the resulting gaps with content that maintains structural similarities to the removed elements. This behavior likely arises from a conflict between CFG—which pushes for object removal—and self-attention guidance—which favors preserving the original structure. Despite this unintended effect, \method{} is the only approach among those compared that actually removes objects, whereas other negative-prompt-based methods fail to do so. Future work may address this by selectively masking self-attention guidance in areas corresponding to the removed elements, similar to the local blending strategy.

\input{fig/ex_structpres.tex}

\subsection{Failure Cases}
\label{sec:sup_limitations_failure}

\input{fig/ex_fail_all.tex}

CFG-based editing methods are inherently sensitive to hyperparameter settings, especially when additional parameters control editing or reconstruction. Figure~\ref{fig:sup_exfailall} presents examples from the PieBench dataset where various methods, including ours, struggle to achieve the intended edits. In many cases, tuning parameters can resolve the issues for SAGE, but such fine-tuning is impractical for large-scale evaluations. Notably, in a small fraction of cases (6 out of 700), SAGE produces oversaturated and poorly reconstructed images—a distinct failure mode compared to other methods. We found that significantly reducing both the CFG weight ($w$) and the self-attention guidance ($\lambda$) effectively mitigates these problems, indicating a promising direction for further refinement.

%% file: tables/comp_time.tex
    \begin{table}[htbp]
    \centering
    \small
      \renewcommand\arraystretch{0.5}
      \setlength{\tabcolsep}{0.5mm}{
  \resizebox{0.9\linewidth}{!}{ 
  \begin{tabular}{ccc}
    \toprule
    \textbf{Method}             &   \textbf{Time} (s)  $\downarrow$  & \textbf{Memory} (GB) $\downarrow$\\ \midrule
    
    
    \nulltextlong{} \cite{mokady2023null} & 115.3	& 21.1\\
    Negative prompt \cite{miyake2023negative} & 26.6	& 38.9\\
    \negproxlong{} \cite{miyake2023negative}  & 23.9 &	38.9 \\
    \textbf{\pnplong{}} \cite{tumanyan2023plug}  & \textbf{12.5}*	& 38.9 \\
    \textbf{\directinversionlong{}} \cite{ju2023direct} & 33.4	& \textbf{12.4}* \\
    \textbf{\directinversionlong{}} \textit{FP16} \cite{ju2023direct} & \textbf{12.5}	& \textbf{7.4} \\
    \ptopzerolong \cite{parmar2023zero} & 52.7 &	25.0\\
    \textbf{\method{}} (ours) & 12.6 & \textbf{7.4} \\
    \method{} \textit{FP32} (ours) & 27.6 &	29.3 \\
    
    \bottomrule
    \end{tabular}
    }

    }
    \caption{
        Performance of prompt-based editing methods on a NVIDIA A100-40GB GPU. The Time column reflects extra time to generate a second 512×512 image (isolating generation cost), and the Memory column shows the peak memory usage per image. FP32 best results are marked with *.
}
            \label{tab:comparison_time}
    \end{table}

%% file: fig/cfg_self_matrix.tex
\begin{figure}[h]
    \centering
    \begin{tabular}{m{1.5cm}m{1.5cm}m{1.5cm}m{0.4cm}m{0.5cm}}

    \includegraphics[width=1.5cm]{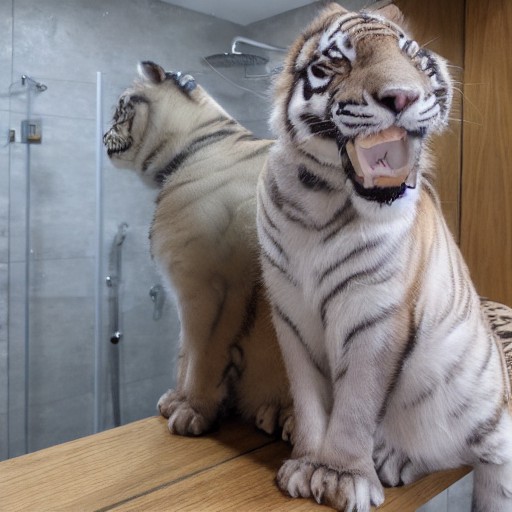} & \includegraphics[width=1.5cm]{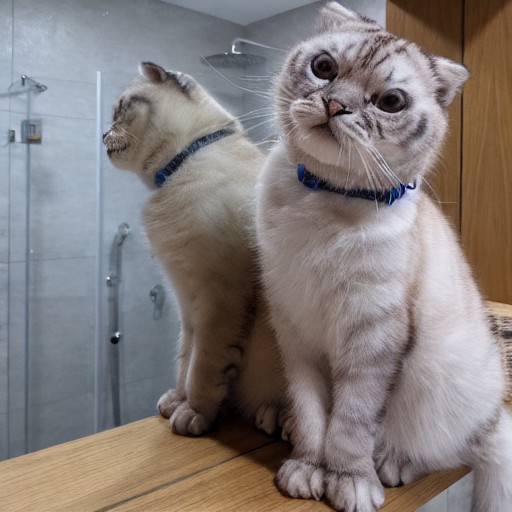} & \includegraphics[width=1.5cm]{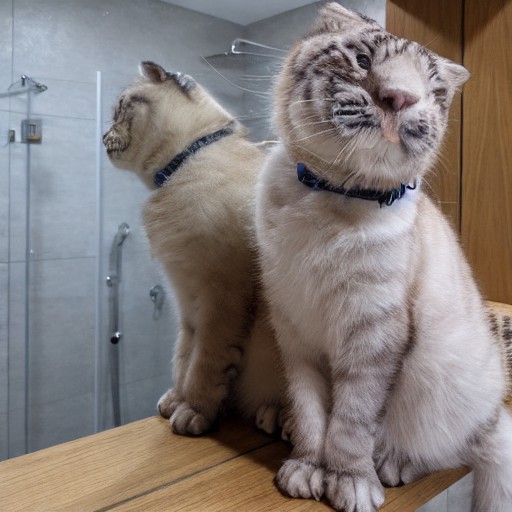} & \rotatebox[origin=c]{90}{$w=2$} & \multirow{3}{*}{\rotatebox[origin=c]{90}{\textit{Classifier-Free guidance}}} \\
    
    \includegraphics[width=1.5cm]{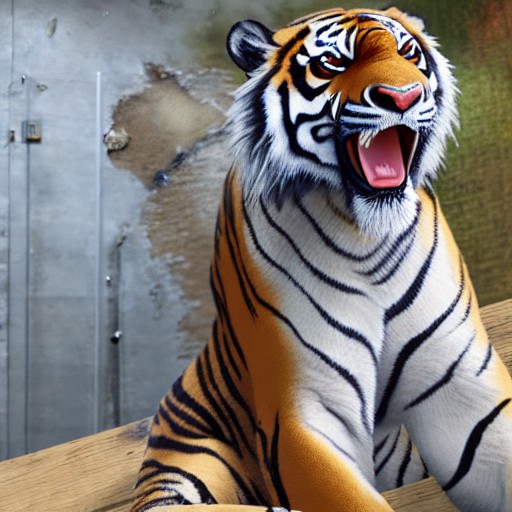} & \includegraphics[width=1.5cm]{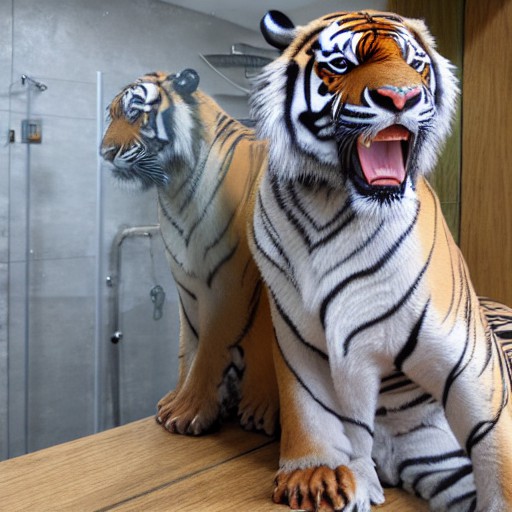} & \includegraphics[width=1.5cm]{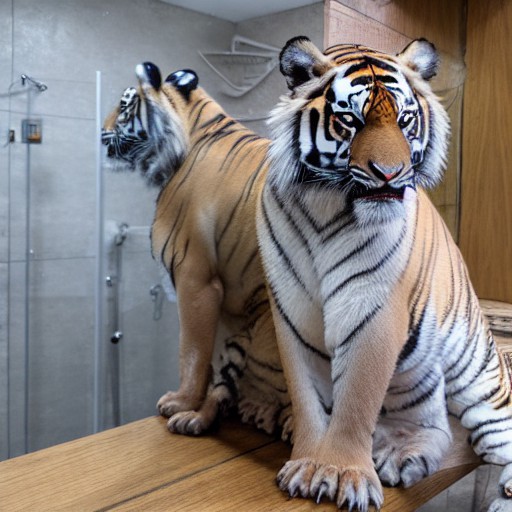} & \rotatebox[origin=c]{90}{$w=7.5$} &  \\
    \includegraphics[width=1.5cm]{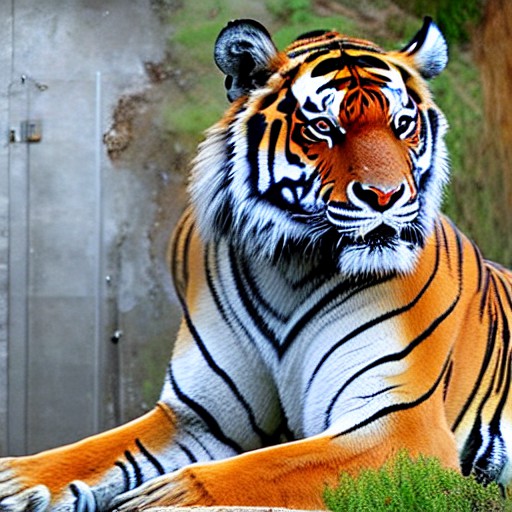} & \includegraphics[width=1.5cm]{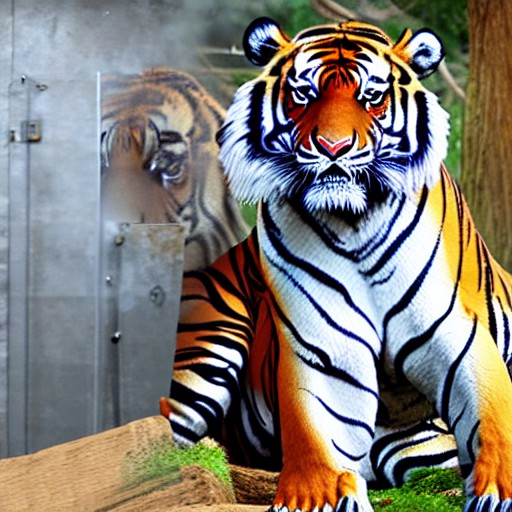} & \includegraphics[width=1.5cm]{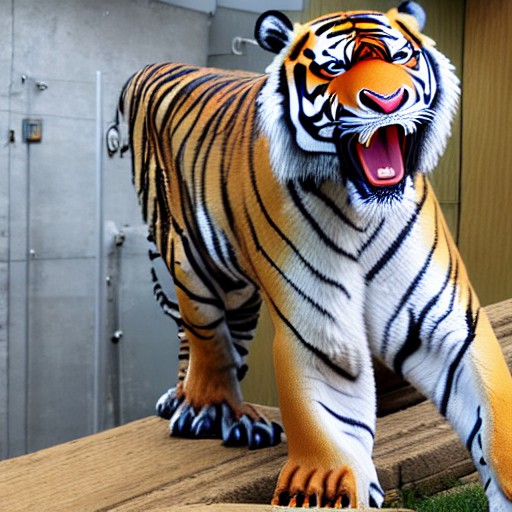} & \rotatebox[origin=c]{90}{$w=15$} &  \\
    \centering  $ \lambda = 50 $  & \centering  $ \lambda = 250 $ & \centering  $ \lambda = 500 $ & & \\
    \multicolumn{3}{c}{\textit{Self-attention guidance}} & & \\
    \end{tabular}
    \caption{This matrix illustrates the interplay between CFG $w$ and self-attention guidance scale $\lambda$, highlighting their influence on image generation. The matrix shows how varying levels of $\lambda$ and $w$ drive the generated image toward either reconstruction (achieved with high $\lambda$ and low $w$) or editing (achieved with low $\lambda$ and high $w$). The images are generated based on the prompt ``'\textit{a \textcolor{BrickRed}{\st{cat}} \textcolor{RoyalBlue}{\textbf{tiger}} sitting next to a mirror}''.}
    \label{fig:supablation_cfg_guidance}
  \end{figure}

%% file: fig/res_reconstruction.tex
\setlength{\tabcolsep}{1pt}
\begin{figure}[h]
    \footnotesize
    \small
    \centering
    \begin{tabularx}{0.7\columnwidth}{ YYYY }
        VAE & \method &  VAE & \method  \\
        \includegraphics[width=\linewidth]{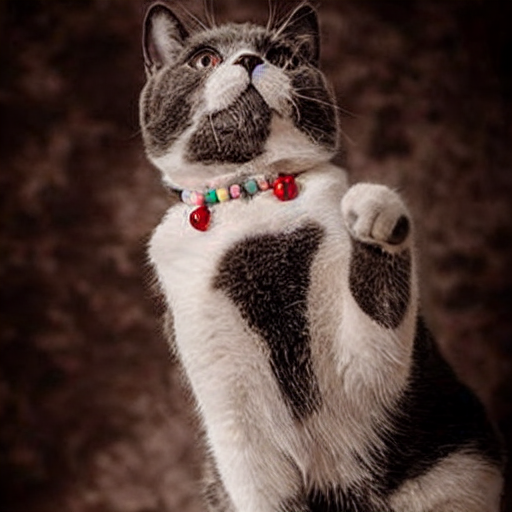} & \includegraphics[width=\linewidth]{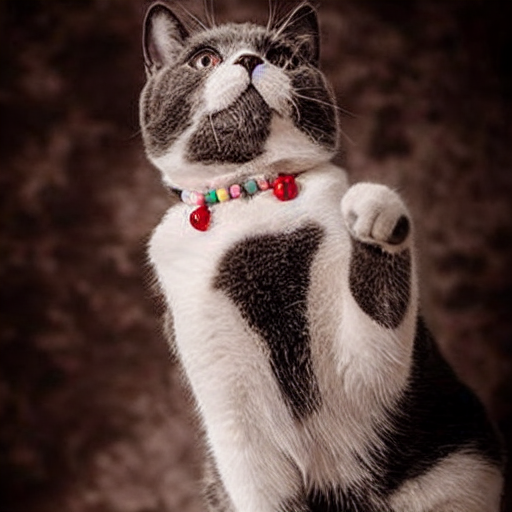} &
        \includegraphics[width=\linewidth]{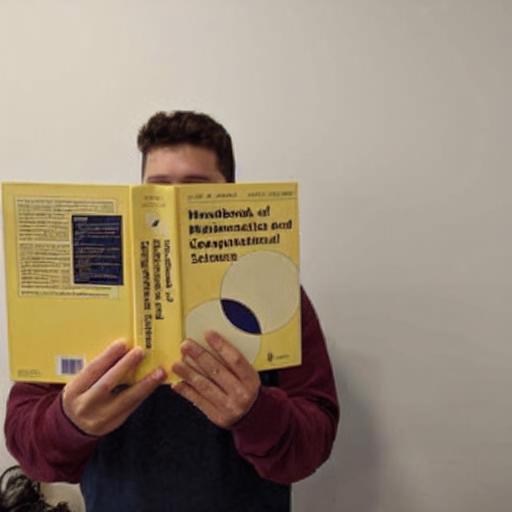} & \includegraphics[width=\linewidth]{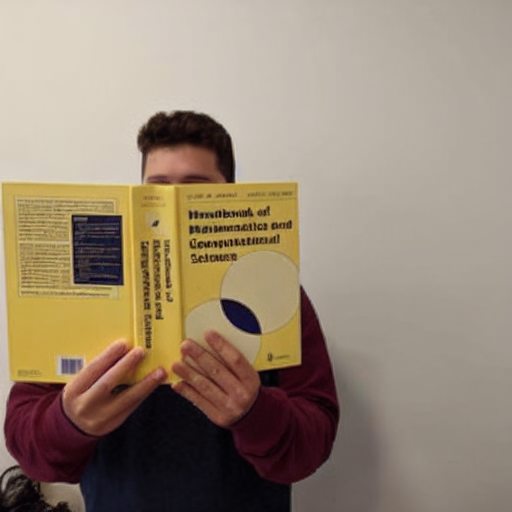}
        \\
        \includegraphics[width=\linewidth]{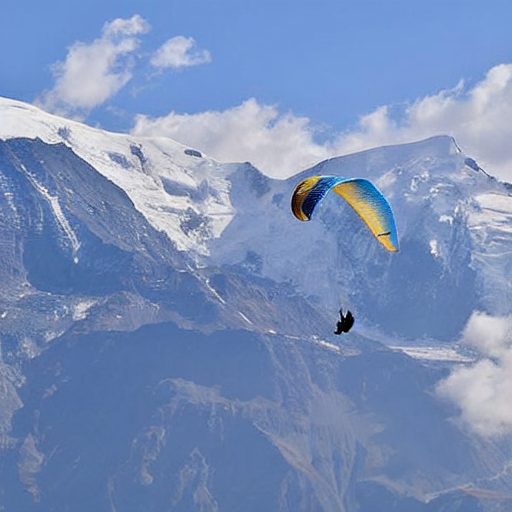} & \includegraphics[width=\linewidth]{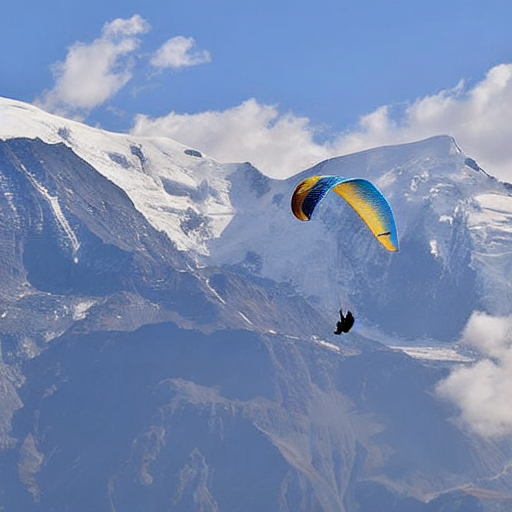} &
        \includegraphics[width=\linewidth]{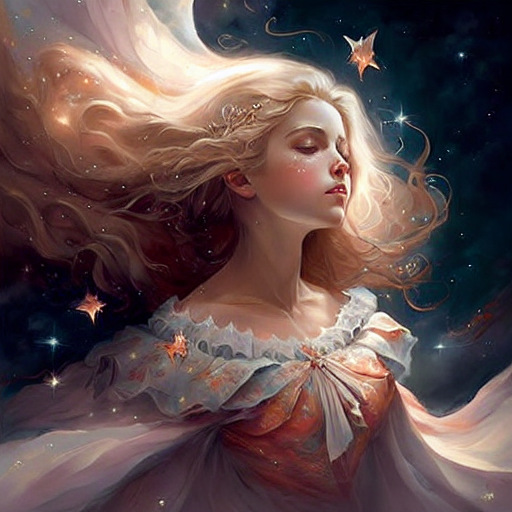} & \includegraphics[width=\linewidth]{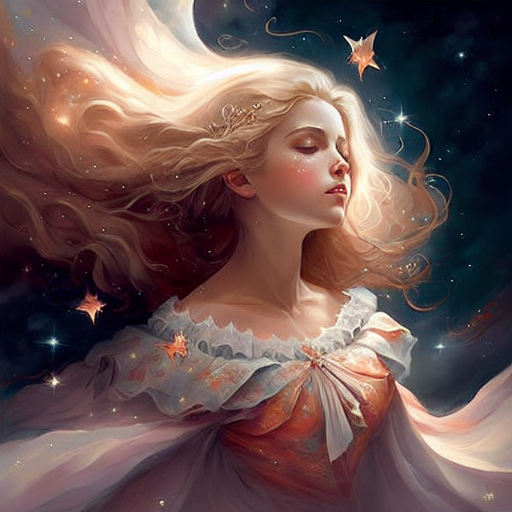}
        \\
        \end{tabularx}
    \caption{Reconstruction performance on PieBench samples.}
    \label{fig:res_reconstruction}
  \end{figure}

%% file: fig/res_reconstruction_table.tex
    \begin{table}[htbp]
        
    \centering
    \small
      \renewcommand\arraystretch{0.9}
      \setlength{\tabcolsep}{0.5mm}{
  \resizebox{\linewidth}{!}{  \begin{tabular}{cccccc}
    \toprule
    \textbf{Method}            & \textbf{Struc. dist.} $\downarrow$ &   \textbf{PSNR} $\uparrow$  & \textbf{LPIPS} $\downarrow$    & \textbf{MSE}  $\downarrow$          & \textbf{SSIM}  $\uparrow$       \\ \midrule
    \textbf{VAE reconstruction} & \textbf{2.8} & \textbf{27.2} & \textbf{40.2} & \textbf{28.5} & \textbf{79.8} \\
    \textbf{\directinversionlong} \cite{ju2023direct} & 3.0 & 27.1 & 51.7 & 28.9 & 79.5\\
    \textbf{\nulltextlong} \cite{parmar2023zero} & 3.3 & 26.7 & 54.8 & 31.1 & 78.9 \\
    \textbf{\method} (ours) & 12.0 & 24.7 & 65.8 & 65.1 & 77.47\\
    
    \bottomrule
    \end{tabular}}

    }
    \caption{
        Performance of different models on PieBench samples, specifically evaluating reconstruction quality only ($\pin$ and $\pout$ are the same).
    }
            
            \label{tab:res_reconstruction}
    \end{table}

%% file: fig/ex_structpres.tex
\setlength{\tabcolsep}{1pt}
\begin{figure}[h]
    \footnotesize
    \small
    \begin{tabularx}{\columnwidth}{ YYYYYY }
        Input & \ptopzero &  \nulltext &  \directinversion &  \negprox &  \method \\
        \multicolumn{6}{c}{\includegraphics[width=\linewidth]{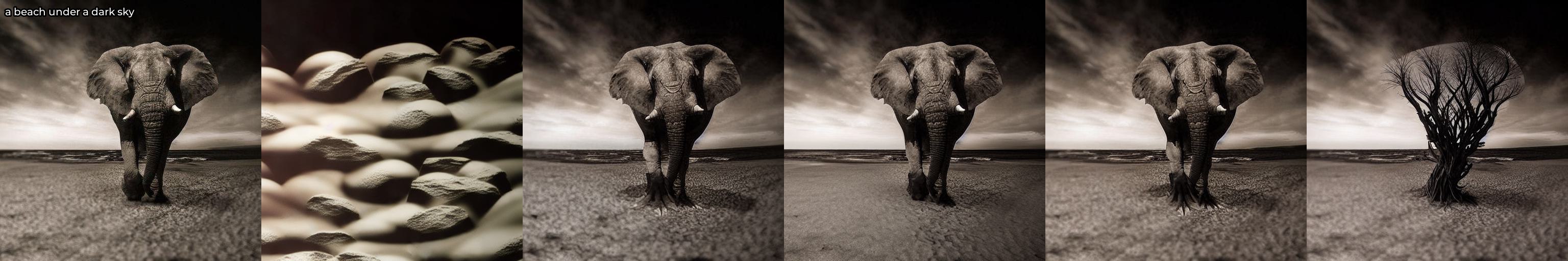}} \\
        \multicolumn{6}{c}{``\textit{\textcolor{BrickRed}{\st{an elephant walking on}} a beach under a dark sky}"} \\
        \multicolumn{6}{c}{\includegraphics[width=\linewidth]{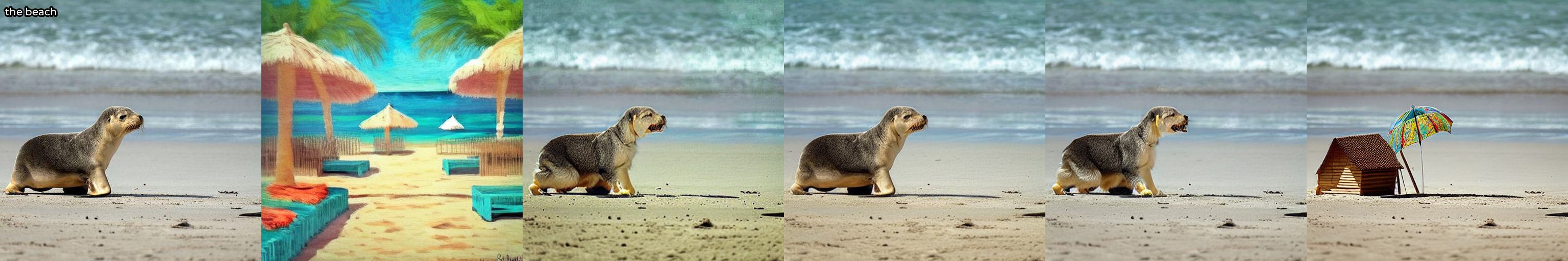}} \\
        \multicolumn{6}{c}{``\textit{\textcolor{BrickRed}{\st{a seal pup on}} the beach}"} \\
        
        \end{tabularx}
    \caption{Examples of object removal by \method{}. Although the method inadvertently fills gaps with structurally similar content, likely due to the complex interplay between CFG and self-attention guidance, it remains the only approach that consistently and effectively removes target objects.}
    \label{fig:sup_structpres}
  \end{figure}

%% file: fig/ex_fail_all.tex
\setlength{\tabcolsep}{1pt}
\begin{figure}[h]
    \footnotesize
    \begin{tabularx}{\columnwidth}{ *{6}{Y} }
        Input & \ptopzero &  \nulltext &  \directinversion &  \negprox &  \method \\
        \multicolumn{6}{c}{\includegraphics[width=\linewidth]{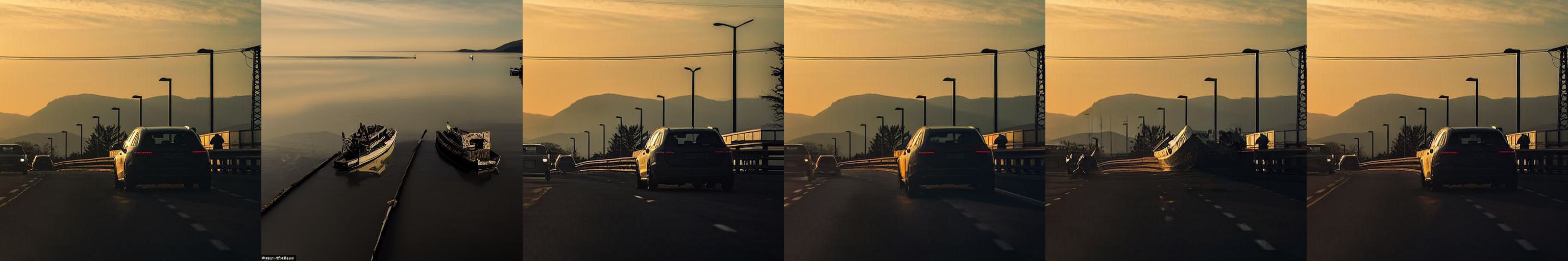}} \\
        \multicolumn{6}{c}{``\textit{\textcolor{BrickRed}{\st{cars}} \textcolor{RoyalBlue}{\textbf{boats}} driving on a highway at sunset}"} \\
        \multicolumn{6}{c}{\includegraphics[width=\linewidth]{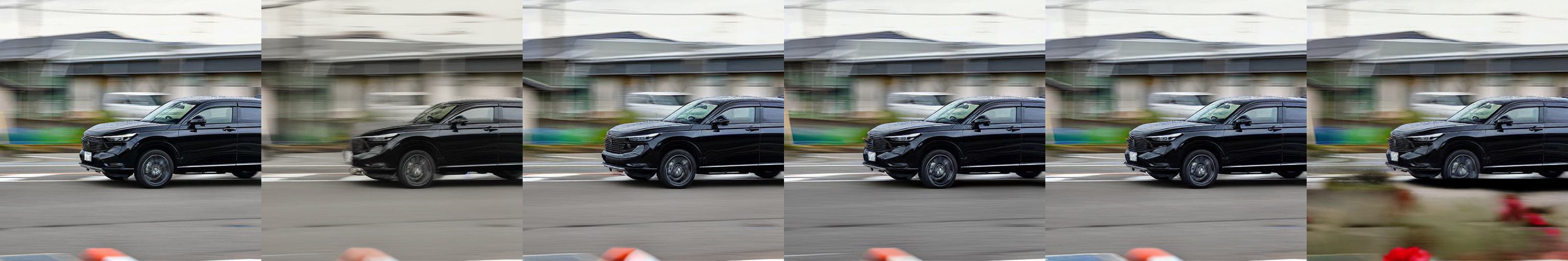}} \\
        \multicolumn{6}{c}{``\textit{the 2020 honda hrx is driving down the road \textcolor{RoyalBlue}{\textbf{full of flowers}}}"} \\
        \multicolumn{6}{c}{\includegraphics[width=\linewidth]{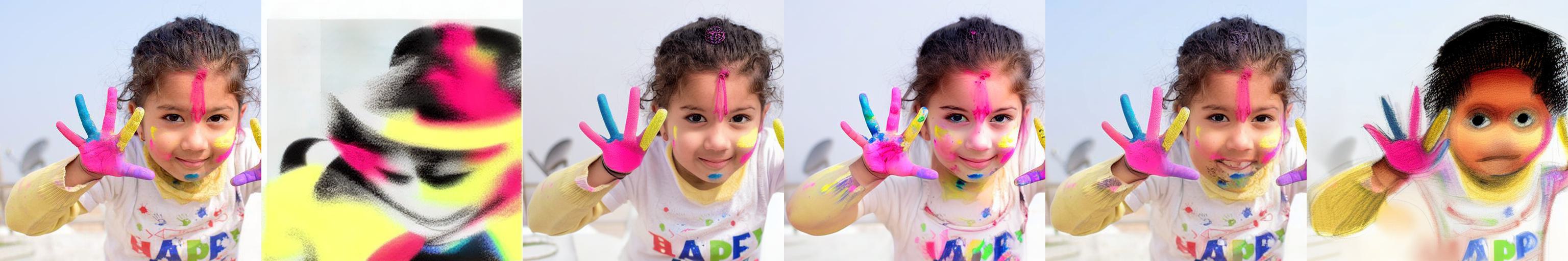}} \\
        \multicolumn{6}{c}{``\textit{\textcolor{RoyalBlue}{\textbf{black and white sketch of}} a young girl with painted hands and face}"} \\
        \multicolumn{6}{c}{\includegraphics[width=\linewidth]{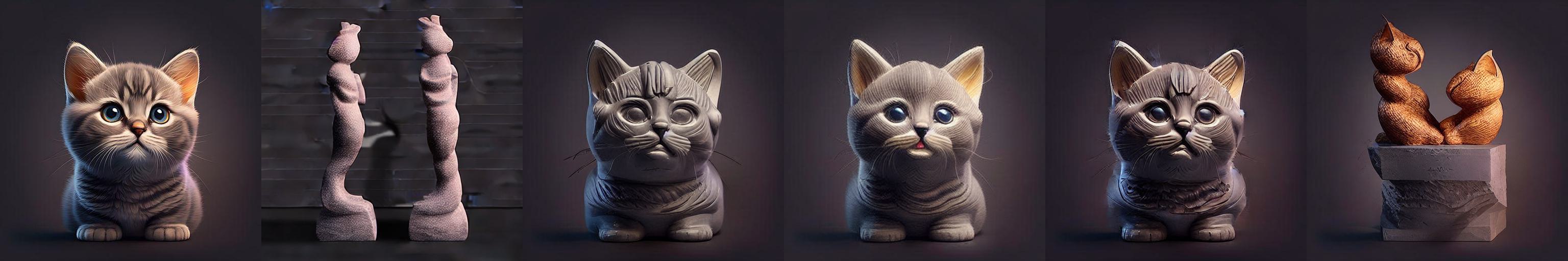}} \\
        \multicolumn{6}{c}{``\textit{a kitten \textcolor{RoyalBlue}{\textbf{sculpture}} }"} \\
        \multicolumn{6}{c}{\includegraphics[width=\linewidth]{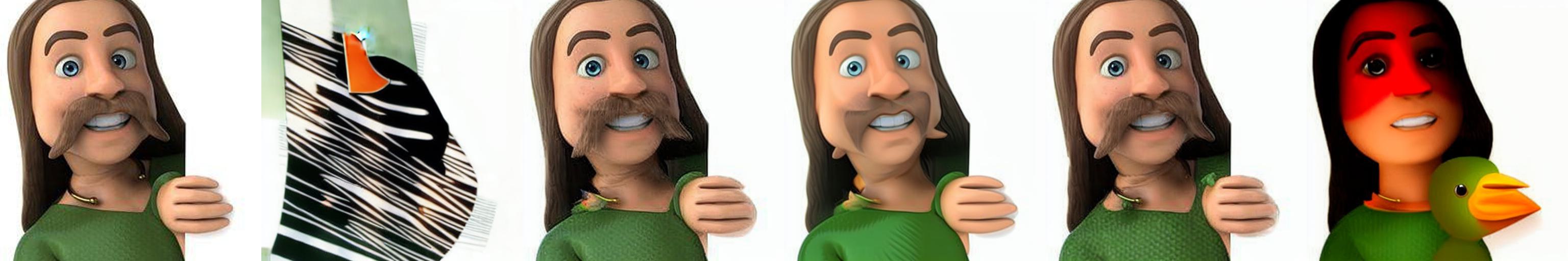}} \\
        \multicolumn{6}{c}{``\textit{a cartoon man \textcolor{RoyalBlue}{\textbf{and a bird}}}"} \\
        \end{tabularx}
        \caption{This figure presents examples from the PieBench dataset where various methods, including ours, encounter difficulties in achieving the desired editing outcomes. These images highlight the challenges and limitations faced in specific editing scenarios, providing insights into areas where each method may require further refinement or adaptation.}
    \label{fig:sup_exfailall}
  \end{figure}